%% file: main.tex
\documentclass[letterpaper]{article} 
\usepackage{aaai2026}  
\usepackage{times}  
\usepackage{helvet}  
\usepackage{courier}  
\usepackage[hyphens]{url}  
\usepackage{graphicx} 
\usepackage{natbib}  
\usepackage{caption} 
\usepackage{algorithm}
\usepackage{algorithmic}

\usepackage{amsmath}
\usepackage{subcaption}
\usepackage{cleveref}
\usepackage{booktabs}
\usepackage{colortbl}
\usepackage{amssymb}
\usepackage{xcolor}
\usepackage{multirow}
\usepackage{multicol}
\usepackage{array}
\usepackage{adjustbox}

\usepackage{newfloat}
\usepackage{listings}
\DeclareCaptionStyle{ruled}{labelfont=normalfont,labelsep=colon,strut=off} 
\floatstyle{ruled}
\newfloat{listing}{tb}{lst}{}
\floatname{listing}{Listing}
\title{Perceive, Act and Correct: Confidence Is Not Enough for Hyperspectral Classification}
\author {
    Muzhou Yang \equalcontrib \textsuperscript{\rm 1}, 
    Wuzhou Quan \equalcontrib \textsuperscript{\rm 1}, 
    Mingqiang Wei \protect\thanks{Mingqiang Wei is the corresponding author.}\textsuperscript{\rm 1}
}
\affiliations {
    \textsuperscript{\rm 1}Nanjing University of Aeronautics and Astronautics\\
}

\usepackage{bibentry}

\begin{document}

\maketitle

\begin{abstract}
Confidence alone is often misleading in hyperspectral image classification, as models tend to mistake high predictive scores for correctness while lacking awareness of uncertainty.
This leads to confirmation bias, especially under sparse annotations or class imbalance, where models overfit confident errors and fail to generalize.
We propose CABIN (Cognitive-Aware Behavior-Informed learNing), a semi-supervised framework that addresses this limitation through a closed-loop learning process of perception, action, and correction.
CABIN first develops perceptual awareness by estimating epistemic uncertainty, identifying ambiguous regions where errors are likely to occur.
It then acts by adopting an Uncertainty-Guided Dual Sampling Strategy, selecting uncertain samples for exploration while anchoring confident ones as stable pseudo-labels to reduce bias.
To correct noisy supervision, CABIN introduces a Fine-Grained Dynamic Assignment Strategy that categorizes pseudo-labeled data into reliable, ambiguous, and noisy subsets, applying tailored losses to enhance generalization.
Experimental results show that a wide range of state-of-the-art methods benefit from the integration of CABIN, with improved labeling efficiency and performance.
\end{abstract}

\begin{links}
    \link{Code}{https://github.com/Muzhou-Yang/CABIN}
\end{links}

\section{Introduction}

\begin{figure}[ht!]
  \centering
    \begin{subfigure}{\linewidth}
      \centering   
      \includegraphics[width=0.98\textwidth]{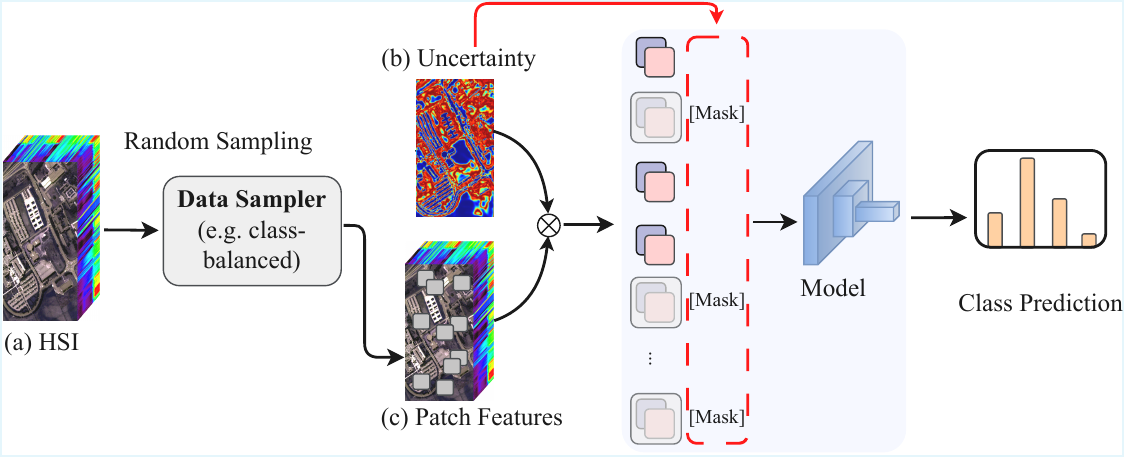}
        \caption{Illustration of data sampling strategies and corresponding uncertainty usage in existing methods.}
        \label{fig:sub1}
    \end{subfigure}
    
    \hfill
    
    \begin{subfigure}{\linewidth}
      \centering   
      \includegraphics[width=0.98\textwidth]{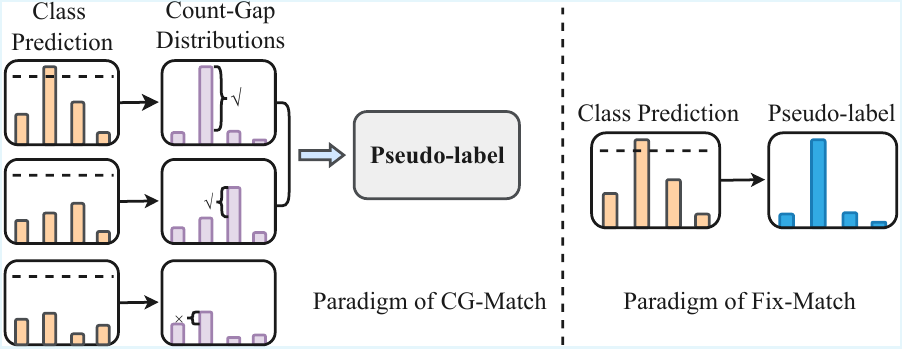}
        \caption{Mainstream semi-supervised methods like FixMatch and CGMatch rely on proxy metrics such as static confidence or past consistency, ignoring the model's current cognitive state.}
        \label{fig:sub2}
    \end{subfigure}
\caption{Existing methods depend on confidence for both sampling and pseudo-labeling, yet overlook cognitive gaps, supporting our core insight that confidence is not enough.}
\label{fig:total}
\end{figure}

Hyperspectral image (HSI) classification enables fine-grained analysis of land cover and material composition by capturing rich spectral signatures across hundreds of contiguous bands.
Its applications are vital in diverse fields such as urban planning~\cite{urban2, Urban}, military reconnaissance~\cite{military}, and precision agriculture~\cite{Crop, crop2}.
Recently, the paradigm of HSI classification has shifted due to the emergence of deep learning techniques.
Advanced methods propose novel mechanisms such as spectral-spatial attention factorization~\cite{SSFTT, SpectralFormer, S3L, GSC_ViT}, self-supervised pre-training~\cite{Scheibenreif_2023_CVPR, FactoFormer}, and state-space sequence modeling~\cite{IGroupSS-Mamba, S2Mamba}, achieving improved accuracy and cross-scene generalization.

Despite these remarkable advances, current methods often rely on the flawed assumption that confident predictions are inherently reliable, even when labels are noisy or ambiguous.
However, due to the limited spatial resolution inherent in HSI data, especially in aerial and satellite imaging, this assumption often fails in practice~\cite{6200362}.
A single pixel may correspond to a mixture of multiple materials or lie on ambiguous boundaries, introducing semantic confusion that reduces the reliability of supervision.
As a result, even carefully annotated ground truth cannot ensure precise or complete information.
In such cases, naive sampling strategies may reinforce confident errors and lead to confirmation bias, where models overfit to seemingly correct annotations that are in fact ambiguous.
This undermines both label efficiency and generalization.

To address this issue, recent studies have turned to uncertainty-aware learning.
Evidential deep learning~\cite{sensoy2018evidentialdeeplearningquantify, Post-hoc_uncertainty} produces principled frameworks to estimate uncertainty for downstream tasks~\cite{EUMS-3D, OpenTAL, FOOD}.
Other methods integrate uncertainty as reference to adjust loss weights or guiding feature suppression~\cite{yu2024uncertaintyaware, SSEL}, as illustrated in Fig.~\ref{fig:sub1}.
However, uncertainty should serve a more active role beyond guiding local optimization or passive learning.
Specifically, it should drive learning behavior: identifying which samples to trust, which regions of the feature space to explore, and where to focus supervision.
Then, we realize this behavior is naturally suitalbe for semi-supervised learning by providing a principled basis for sample selection.
Existing semi-supervised methods such as FixMatch~\cite{sohn2020fixmatchsimplifyingsemisupervisedlearning} adopt fixed confidence thresholds to generate pseudo-labels, and CGMatch~\cite{cgmatch} improves upon this by incorporating historical label consistency.
Despite their strengths, these methods still depend on fixed rules or outdated feedback, making them slow to adapt to the model’s evolving uncertainty during training, as illustrated in Fig.~\ref{fig:sub2}.

Inspired by these insights, we propose \textit{Cognitive-Aware Behavior-Informed learNing} (\textbf{CABIN}), a semi-supervised framework that positions uncertainty as an active driver of both data selection and behavior correction.
CABIN establishes a closed-loop learning process of perception, action, and correction, where the model learns to perceive sample uncertainty, take informed sampling actions, and iteratively correct its own supervision.
This loop is anchored by the \textit{Uncertainty-Guided Dual Sampling Strategy} (UGDSS), which dynamically balances the exploration of uncertain feature regions and the reinforcement of confident predictions.
It encourages the model to discover complex patterns while maintaining stability through reliable supervision.
To further address the issue of unreliable supervision, CABIN introduces a novel criterion, the Uncertainty-Gap metric ($UG_\alpha$), that quantifies the discrepancy between behavioral confidence and epistemic evidence.
Based on $UG_\alpha$, we develop the \textit{Fine-Grained Dynamic Assignment Strategy} (FDAS), which categorizes pseudo-labeled data into reliable, ambiguous, and noisy subsets, applying targeted training objectives accordingly.
This fine-grained supervision enhances robustness under class imbalance, sparse labels, and distribution shifts.
Experimental results demonstrate that integrating CABIN into state-of-the-art methods enables them to achieve superior performance under semi-supervised settings (using only half of the original training labels) with virtually no additional computational costs.

Our contributions can be summarized as follows:
\begin{itemize}
    \item We revisit low-supervision hyperspectral classification and find that confidence alone cannot reliably guide learning. To tackle this, we propose CABIN, an uncertainty-guided framework forming a closed loop of perception, action, and correction.
    
    \item We propose UGDSS, a dual sampling strategy that leverages epistemic uncertainty to balance the exploration of ambiguous regions and the exploitation of reliable pseudo-labels.

    \item We introduce FDAS, a dynamic assignment mechanism that uses the Uncertainty-Gap metric to correct pseudo-label noise through fine-grained supervision.
    
    \item Experiments confirm that CABIN, as a model-agnostic and plug-and-play strategy, can significantly improve the performance of various state-of-the-art methods under semi-supervised settings, even with as little as 75\% of the original annotations.
\end{itemize}

\section{Method}

\subsection{Framework Overview}

\begin{figure*}[ht]
    \centering
    \includegraphics[width=0.95\textwidth]{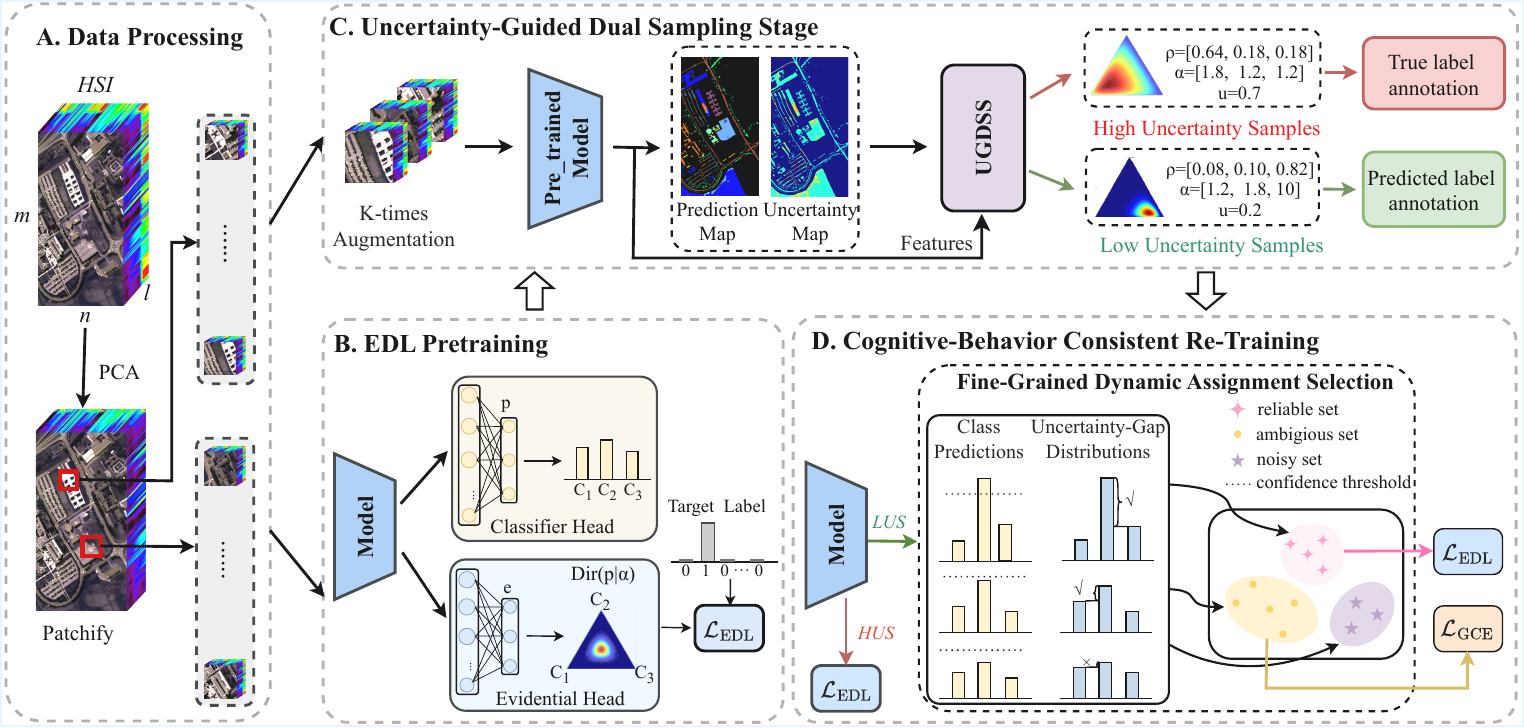}
    \caption{
        An overview of the proposed CABIN framework, designed to address the challenge that ``confidence is not enough" for HSI classification. 
        CABIN establishes a closed-loop process of perception, action and correction: the model first perceives sample uncertainty using EDL, then acts by selecting key samples via UDGSS, and finally corrects cognitive-behavioral gaps by dynamically assessing pseudo-label reliability with FDAS. 
        This process enables more robust learning under limited supervision.
}
    \label{fig:framework}
\end{figure*}

In this section, we revisit the cognitive mechanism of EDL, and propose a novel framework \textbf{CABIN} which explicitly models a cognitive loop of perception, action, and correction to address the core challenge identified in this work: \textit{confidence is not enough} for reliable decision making in HSI classification. 
The full pipeline is illustrated in Fig.~\ref{fig:framework}.

Given a sample $i$, let $\mathbf{e}_i = f(\mathbf{x}_i \mid \boldsymbol{\Theta})$ represent the non-negative evidence vector predicted by the model with parameters $\boldsymbol{\Theta}$ and features $\mathbf{x}_i$.
The corresponding Dirichlet distribution is parameterized as $\boldsymbol{\alpha}_i = \mathbf{e}_i + 1$, with total evidence $S_i = \sum_{k=1}^{K} \alpha_{i,k}$.
The estimated class probabilities are computed as the mean of the Dirichlet distribution: $\hat{\mathbf{p}}_i = \boldsymbol{\alpha}_i / S_i$.
The epistemic uncertainty is denoted as $u_i = K / S_i$.
This perceived uncertainty then guides the model to act through our UGDSS module.
UGDSS partitions the candidate set $\mathcal{D}_t$ based on an adaptive epistemic threshold:
\begin{equation}
\mathcal{D}_{hc}, \mathcal{D}_{qu} = \text{UGDSS}(\mathcal{D}_t, \{u_i\}_{ i \in \mathcal{D}_t}),
\end{equation}
where $\mathcal{D}_{hc}$ comprises high uncertain samples for critical learning, and $\mathcal{D}_{qu}$ includes reliable samples to maintain stability.
To enhance the representation of $\mathcal{D}_{au}$, UGDSS integrates two internal components: (i) Diverse-Representative Query Selection (DRQS) refines $\mathcal{D}_{hc}$ via feature-space clustering:
$
\mathcal{D}_{au} = \text{DRQS}(\mathcal{D}_{hc}). 
$
(ii) Gaussian Feature Perturbation (GFP) generates local variants of selected samples by injecting uncertainty-scaled noise in the feature space:
$
\widehat{\mathcal{D}}_{\text{aug}} = \text{GFP}(\mathcal{D}_{au}).
$
Given the potential noise in the pseudo-labeled set $D_{qu}$, we introduce Fine-Grained Dynamic Assignment Strategy (FDAS), which jointly considers softmax confidence and evidential separation to categorize:
\begin{equation}
    \mathcal{D}_{re}, \mathcal{D}_{am}, \mathcal{D}_{no} = \text{FDAS}(\mathcal{D}_{qu}),
\end{equation}
where $\mathcal{D}_{re}$ contains reliable pseudo-labels, $\mathcal{D}_{am}$ includes ambiguous ones, and $\mathcal{D}_{no}$ is discarded.

The final training objective integrates all confident supervision sources.
For labeled and highly reliable pseudo-labeled samples, we adopt the EDL loss:
\begin{equation}
\mathcal{L}_{\text{EDL}}^{(i)} = \sum_{j=1}^K y_{i,j} \left( \log(S_i) - \log(\alpha_{i,j}) \right),
\end{equation}
where $\mathbf{y}_i \in \{0,1\}^K$ is the one-hot supervision vector.

For ambiguous pseudo-labeled samples, we apply the noise-robust Generalized Cross Entropy (GCE) loss~\cite{zhang2018generalizedcrossentropyloss}:
\begin{equation}
\begin{aligned}
\mathcal{L}_{\text{GCE}}^{(i)} &= \frac{1 - (\hat{p}_{i,y})^q}{q}, \\
\hat{p}_{i,y} &= \frac{\alpha_{i,y}}{S_i},
\end{aligned}
\end{equation}
where \( y \) is the pseudo-label index for sample \( i \), and the hyperparameter \( q \in (0, 1] \) is set to 0.7 following~\cite{zhang2018generalizedcrossentropyloss}.

The total training objective is:
{\small
\begin{equation}
\mathcal{L} = \mathcal{L}_{\text{EDL}}(\mathcal{D}_L \cup \widehat{\mathcal{D}}_{\text{aug}}) + \lambda_{r} \mathcal{L}_{\text{EDL}}(\mathcal{D}_{re}) + \lambda_{a} \mathcal{L}_{\text{GCE}}(\mathcal{D}_{am}),
\end{equation}}
where $\lambda_r$ and $\lambda_a$ are balancing weights, both set to 0.3.

\subsection{Uncertainty-Guided Dual-Sampling Strategy}

\begin{figure*}[t]
    \centering
    \includegraphics[width=0.97\linewidth]{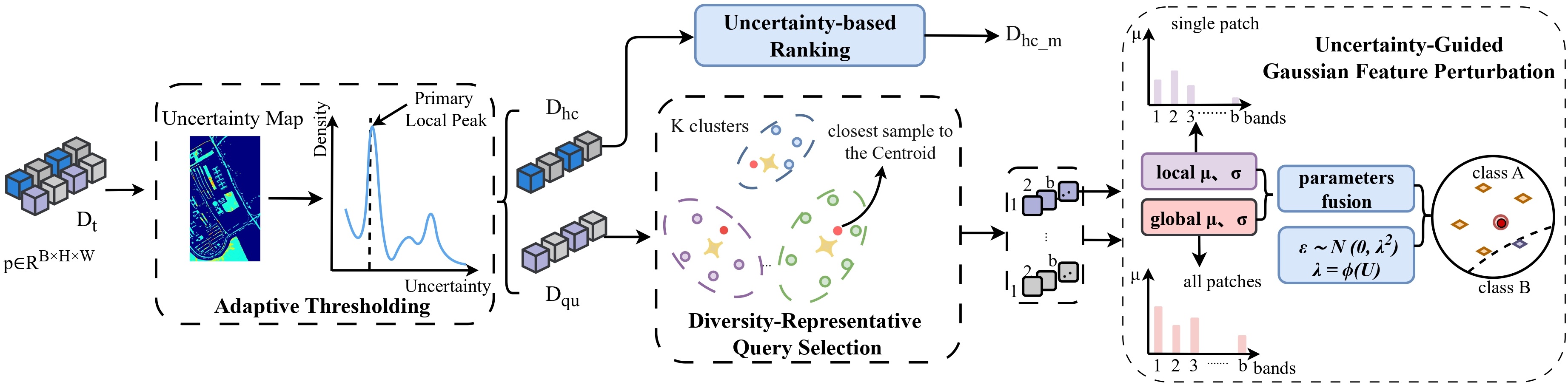}
    \caption{
        Schematic of the UGDSS module, which leverages epistemic uncertainty to orchestrate a dual-path selection strategy balancing exploration and exploitation.
        The exploration path combines DRQS and GFP to actively probe the model's knowledge gaps, whereas the exploitation path selects confident samples for reliable pseudo-labeling.
}
    \label{fig:UGDSS}
\end{figure*}

Building upon EDL, we propose UGDSS (in Fig.~\ref{fig:UGDSS}) which partitions the candidate set $\mathcal{D}_t$ into two subsets for targeted exploration and exploitation. 
Since EDL-based uncertainty estimation is particularly sensitive to disturbances, which may lead to a misleading sampling process especially under sparse data conditions, we adopt a test-time augmentation method~\cite{TTA}, generating $K$ spectral-spatial variants of each $\mathbf{x}_i$ through $k$-th random transforms $\mathcal{T}_k$, and averaging their EDL outputs:
\begin{equation}
    \hat{u}_i=\frac{1}{K}\sum_{k=1}^{K}u(\mathcal{T}_k(\mathbf{x}_i)).
\end{equation}

\subsubsection{Adaptive Thresholding.}
The estimated uncertainty values $\{\hat{u}_i\}_{i=1}^{N}$ exhibit a heterogeneous distribution, reflecting epistemic diversity across the feature space.
However, using a static cutoff (e.g., fixed percentile) may introduce misaligned partitions due to the early-stage model’s prediction noise and instability.

To this end, we introduce a histogram-based adaptive thresholding strategy within the UGDSS module. Specifically, we sort $\hat{u}_i$ in ascending order to form a histogram density distribution $h[n]$ with $N$ bins.
The adaptive threshold $T_u \in \mathbb{R}$ is set as the first local minimum satisfying:
\begin{equation}
    T_u = \min \left\{ h[n] \,\middle|\, \Delta h[n] < \delta,\; \Delta^2 h[n] < 0 \right\},
\end{equation}
where $\Delta h[n] = h[n{+}1] - h[n]$ and $\Delta^2 h[n] = \Delta h[n{+}1] - \Delta h[n]$ denote the first- and second-order discrete differences. The tolerance $\delta$ avoids unstable extrema.

This adaptive threshold $T_u$ is then used to divide the current candidate set $\mathcal{D}_t$ into two subsets:
$
\mathcal{D}_{hc} = \left\{ i \in \mathcal{D}_t \;\middle|\; \hat{u}_i \ge T_u \right\}, \quad \mathcal{D}_{qu} = \mathcal{D}_t \setminus \mathcal{D}_{hc}. 
$
This enables UGDSS to adaptively partition samples based on current uncertainty distribution, facilitating the robust selection of both uncertain and reliable samples.

\subsubsection{Diverse-Representative Query Selection.}
To further refine the high-uncertainty candidate set $\mathcal{D}_{hc}$ obtained via UGDSS, we propose DRQS strategy to eliminate sample redundancy and enhance selection diversity.

Specifically, for each candidate sample $i \in \mathcal{D}_{hc}$, we extract its semantic embedding $\bar{f}_i$ from the final feature layer of the model $\Phi$, which captures high-level discriminative information prior to classification.
We then perform K-means++~\cite{kmeans} clustering over the embedding set $\{ \bar{f}_i \}$ and partition it into $M$ clusters, where $M$ is the predefined query budget.
From each cluster, the sample closest to the centroid is selected to form the final query set $\mathcal{D}_{au}$:
\begin{equation}
\mathcal{D}_{au} = \bigcup_{k=1}^{M} \left\{ \arg\min_{i \in \mathcal{D}_{hc}} \left\| \bar{f}_i - C_k \right\|_2 \right\},
\end{equation}
where $C_k$ denotes the centroid of the $k$-th cluster, and $\| \cdot \|_2$ represents the Euclidean distance. 

\subsubsection{Uncertainty-Guided Gaussian Feature Perturbation.}
To enhance the model’s robustness in highly uncertain regions, we introduce an uncertainty-guided Gaussian perturbation module that promotes feature-level exploration.
The core idea is to make the perturbation strength proportional to the model’s uncertainty.
For each sample $i \in \mathcal{D}_{au}$, we first compute its perturbation scale $\lambda_i$ by linearly mapping the normalized epistemic uncertainty $\hat{u}_i \in [0, 1]$ to a predefined range $[\lambda_{\min}, \lambda_{\max}]$:
\begin{equation}
\lambda_i = \lambda_{\min} + (\lambda_{\max} - \lambda_{\min}) \cdot \hat{u}_i.
\end{equation}

This ensures that higher uncertainty yields stronger perturbations.
We then generate a perturbed feature representation by injecting Gaussian noise scaled by $\lambda_i$.
The noise is sampled from a weighted mixture of local and global statistics over $\mathcal{D}_{au}$.
To maintain spectral fidelity and prevent unrealistic distortions, all perturbed values are clamped to their original dynamic range.
This feature-space augmentation strategy, inspired by methods like Manifold Mixup~\cite{Manifold_Mixup} and CutMix~\cite{CutMix}, encourages the model to learn more invariant features.
The resulting augmented set, denoted as $\hat{\mathcal{D}}_{aug}$, is then used for retraining.

\begin{table*}[t]
    \centering
    \small
    \setlength{\tabcolsep}{6.4pt}
    \begin{tabular*}{\linewidth}{c||ccc||c>{\columncolor{gray!30}}cc>{\columncolor{gray!30}}c||c>{\columncolor{gray!30}}cc>{\columncolor{gray!30}}c}
        \toprule
        & \multicolumn{3}{c||}{}
        & \multicolumn{4}{c||}{\textbf{CNN-based}}
        & \multicolumn{4}{c}{\textbf{Transformer-based}}
        \\ \midrule
        Class No.
        & \multicolumn{1}{c}{SVM} 
        & \multicolumn{1}{c}{2DCNN} 
        & \multicolumn{1}{c||}{3DCNN} 
        & \multicolumn{1}{c}{$\mathrm {ReS^2}$} 
        & \multicolumn{1}{>{\columncolor{gray!30}}c}{CABIN} 
        & \multicolumn{1}{c}{CLOLN} 
        & \multicolumn{1}{>{\columncolor{gray!30}}c||}{CABIN} 
        & \multicolumn{1}{c}{SSFTT} 
        & \multicolumn{1}{>{\columncolor{gray!30}}c}{CABIN} 
        & \multicolumn{1}{c}{GSC-ViT} 
        & \multicolumn{1}{>{\columncolor{gray!30}}c}{CABIN} \\

        \midrule
        \midrule
        1 & 19.76 & 65.63 & 75.12 & 100.00 & \textbf{100.00} & 100.00 & \textbf{100.00}  & 100.00 & \textbf{100.00} & 100.00 & \textbf{100.00} \\ 
        2 & 71.60 & 63.44 & 70.40 & 48.63 & \textbf{68.08} & 64.91 & \textbf{77.95} & \textbf{88.04} & 85.23 & 59.08 & \textbf{86.17}\\ 
        3 & 41.90 & 60.25 & 68.43 & 71.14 & \textbf{76.96} & 61.01 & \textbf{74.30}  & 80.13 & \textbf{82.28} & \textbf{82.91} &  76.20\\ 
        4 & 79.81 & 41.11 & 69.73 & 99.49 & \textbf{100.00} & 91.37 & \textbf{100.00} & 95.94 & \textbf{100.00} & \textbf{100.00} & 96.45\\ 
        5 & 25.35 & 87.05 & 77.16 & 88.26 & \textbf{96.16} & 90.97 & \textbf{91.87} & 87.58 & \textbf{93.45} & 84.88 & \textbf{94.58} \\ 
        6 & 86.30 & 97.21 & 85.70 & 95.0 & \textbf{98.12} & 95.65  & \textbf{95.94} & \textbf{96.52} & 94.93 & 91.01 & \textbf{99.13}\\ 
        7 & 34.00 & 89.47 & 70.63 & 100.00 & \textbf{100.00} & 100.00 & \textbf{100.00} & 100.00&  \textbf{100.00} & 100.00 & \textbf{100.00} \\ 
        8 & 95.12 & 96.66 & 90.21 & 100.00 & \textbf{100.00} & 100.00 & \textbf{100.00} & 100.00 & \textbf{100.00} & \textbf{100.00} & 96.12 \\ 
        9 & 44.44 & 32.26 & 68.20 & 100 & \textbf{100.00} & 100.00 & \textbf{100.00} & 100.00 & \textbf{100.00} & 100.00 & \textbf{100.00}\\
        10 & 52.75 & 73.84 & 75.48 & 62.98 & \textbf{96.35} & \textbf{90.56} & 83.05 & 75.86 & \textbf{92.49} & \textbf{91.20} & 89.06\\ 
        11 & 87.19 & 84.36 & 78.57 & 81.16 & \textbf{81.99} & \textbf{80.79} & 79.38 & 85.05 & \textbf{87.00} & \textbf{84.43} & 78.88\\ 
        12 & 42.78 & 42.89 & 65.23 & 83.54 & \textbf{83.73} & \textbf{88.79} & 66.00 & 74.14 & \textbf{77.22} & 80.83 & \textbf{89.69}\\
        13 & 92.39 & 98.58 & 90.17 & 100.00 & \textbf{100.00} & \textbf{100.00} & 99.39 & 100 & \textbf{100.00} & 99.39 & \textbf{100.00}\\
        14 & 96.84 & 94.02 & 94.65 & 98.04 & \textbf{99.92} & \textbf{98.12} & 95.84 & 94.37 & \textbf{97.88} & \textbf{95.18} & 91.10\\
        15 & 63.40 & 42.65 & 64.89& 95.38 & \textbf{99.42} & 96.23 & \textbf{97.69} & \textbf{99.71} & 97.40 & 92.49 & \textbf{99.42}\\
        16 & 90.36 & 92.19 & 93.07 & 100.00 & \textbf{100.00} & 98.11 & \textbf{100.00} & 100.00 & \textbf{100.00} & \textbf{100.00} & 96.23\\ \midrule

        \multirow{2}{*}{OA(\%)} & \multirow{2}{*}{72.51} & \multirow{2}{*}{76.39} & \multirow{2}{*}{78.28} & \multirow{2}{*}{79.70} & \textbf{87.39} & \multirow{2}{*}{83.77}  & \textbf{84.69} & \multirow{2}{*}{87.35} & \textbf{90.08} & \multirow{2}{*}{84.64} & \textbf{87.40}\\ 
         &       &       &       &       & \textcolor[rgb]{0.0, 0.6, 0.0}{+7.69} &       & \textcolor[rgb]{0.0, 0.705, 0.0}{+0.92} &       & \textcolor[rgb]{0.0, 0.6, 0.0}{+2.73} &  & \textcolor[rgb]{0.0, 0.6, 0.0}{+2.76} \\
        \multirow{2}{*}{AA(\%)} & \multirow{2}{*}{65.50} & \multirow{2}{*}{72.18} & \multirow{2}{*}{76.94} & \multirow{2}{*}{88.98} & \textbf{93.80} & \multirow{2}{*}{90.34} & \textbf{91.22} & \multirow{2}{*}{93.65} & \textbf{94.20} & \multirow{2}{*}{91.33} & \textbf{93.31}\\
        &  &  &  & & \textcolor[rgb]{0.0, 0.6, 0.0}{+4.82} & & \textcolor[rgb]{0.0, 0.6, 0.0}{+0.88} & & \textcolor[rgb]{0.0, 0.6, 0.0}{+0.55} &  & \textcolor[rgb]{0.0, 0.6, 0.0}{+1.98} \\
        \multirow{2}{*}{$\kappa \times 100$} & \multirow{2}{*}{68.22} & \multirow{2}{*}{72.85} & \multirow{2}{*}{75.46} & \multirow{2}{*}{76.71} & \textbf{85.68} & \multirow{2}{*}{81.40} & \textbf{82.60} & \multirow{2}{*}{85.48} & \textbf{88.66} & \multirow{2}{*}{82.49} & \textbf{85.60}\\
        &  &  &  & & \textcolor[rgb]{0.0, 0.6, 0.0}{+8.97} & & \textcolor[rgb]{0.0, 0.6, 0.0}{+1.20} & & \textcolor[rgb]{0.0, 0.6, 0.0}{+3.18} &  & \textcolor[rgb]{0.0, 0.6, 0.0}{+3.11} \\
        \bottomrule

    \end{tabular*}
    \caption{
        Comparison of classification and overall performance on the \textit{Indian Pines} dataset across different methods.
        For other methods, 20 samples per class (320 total) are used, while CABIN uses only 240 samples.
        CABIN results are highlighted in light gray.
        Red font indicates performance degradation, while Green font indicates improvement.
        The best results for both per-class accuracy and overall metrics, comparing CABIN and non-CABIN methods, are shown in \textbf{bold}.
        The same settings are applied in the next table.}
    \label{tab:constract_experiments}
\end{table*}

\subsection{Fine-Grained Dynamic Assignment Strategy}
Existing methods like CGMatch~\cite{cgmatch} identify ambiguous samples by tracking historical pseudo-label distributions using the Count-Gap metric.
However, this method incurs delayed responses, high computational cost, and poorly reflects the model's current confidence.
To address these limitations, we propose a cognitive-driven pseudo-label selection framework named FDAS.
By jointly leveraging the behavioral confidence and internal evidential cues, FDAS enables a dynamic separation of the pseudo-labeled set $\mathcal{D}_{pu}$.

Specifically, behavioral confidence is denoted as the maximum classification probability $c_i=\max(\mathbf{\hat{p}}_i)$, where $\mathbf{\hat{p}}_i$ denotes the predicted class distribution for sample $i \in \mathcal{D}_{pu}$.
To capture the model's discriminative certainty from a cognitive perspective, we introduce the \textit{Uncertainty-Gap} ($UG_\alpha$) based on EDL.
Given the evidence vector $\boldsymbol{\alpha}_i = (\alpha_{i1}, \dots, \alpha_{iK})$ for a $K$-class sample $x_i$, we apply an Exponential Moving Average (EMA) to obtain a smoothed vector $\bar{\boldsymbol{\alpha}}_i$, thereby mitigating fluctuations from single-batch estimations:
\begin{equation}
    \bar{\boldsymbol{\alpha}}_i \leftarrow m \cdot \bar{\boldsymbol{\alpha}}_i + (1 - m) \cdot \boldsymbol{\alpha}_i,
\end{equation}
where $m\in[0,1)$ is the momentum coefficient. The UG is then computed using this smoothed evidence:
$UG_i^\alpha = \max_k(\bar{\alpha}_{ik}) - \text{second\_max}_k(\bar{\alpha}_{ik})$,
which quantifies the evidential margin between the most and second-most supported classes.

Based on this stable metric, we partition the pseudo-labeled set $\mathcal{D}_{pu}$ into three disjoint subsets using dynamic thresholds for confidence ($\tau_c$) and evidence ($\tau_e$):
\begin{equation}
\begin{aligned}
    \mathcal{D}_{re} &= \left\{x_i \in \mathcal{D}_{pu} \ \middle| \ c_i \ge \tau_c \  \& \ \text{UG}_i^\alpha \ge \tau_e \right\}, \\
    \mathcal{D}_{no} &= \left\{x_i \in \mathcal{D}_{pu} \ \middle| \ c_i < \tau_c \  \& \ \text{UG}_i^\alpha < \tau_e \right\}, \\
    \mathcal{D}_{am} &= \mathcal{D}_{pu} \setminus (\mathcal{D}_{re} \cup \mathcal{D}_{no}),
\end{aligned}
\end{equation}
where $\mathcal{D}_{re}$, $\mathcal{D}_{no}$, and $\mathcal{D}_{am}$ correspond to the reliable, noisy, and ambiguous sets, respectively.
This three-way partitioning allows for tailored learning strategies for each subset.
Moreover, both thresholds $\tau_c$ and $\tau_e$ are also dynamically updated using EMA over their respective batch-wise statistics (i.e., the average confidence and the average uncertainty gap), allowing the model to adapt to its evolving confidence.

\section{Experiments}

\subsection{Datsets and Evaluation Metrics} 
We conduct experiments on five widely used hyperspectral datasets, including Indian Pines, Salinas, Pavia University, WHU-Hi-HongHu, and WHU-Hi-LongKou~\cite{whuhi}.
These datasets vary in spatial resolution, spectral range, and scene complexity, covering agricultural, urban, and UAV-based remote sensing scenarios.
Overall, they provide a comprehensive benchmark for evaluating robust hyperspectral classification methods.


Based on the above datasets, we evaluate the performance of our method using several standard classification metrics, including Overall Accuracy (OA), Average Accuracy (AA), and Cohen’s Kappa Coefficient~\cite{k} ($\kappa$). 

\subsection{Comparative Experiments}

To validate the effectiveness, we integrate CABIN into four existing state-of-the-art methods, including CNN-based (ReS$^2$~\cite{ReS2}, CLOLN~\cite{Li_CLOLN}) and Transformer-based (SSFTT~\cite{SSFTT}, GSC-ViT~\cite{GSC_ViT}).
In addition, three representative baseline methods are included for comprehensive comparison.
Model architectures follow the original implementations to ensure fair comparison, while training hyperparameters are adjusted to fit our experimental setup.

Overall, CABIN delivers substantial performance gains across nearly all models and datasets.
Especially on Indian Pines (Table~\ref{tab:constract_experiments}), it improves ReS$^2$'s OA by a remarkable \textbf{+7.69\%} and SSFTT's $\kappa$ by \textbf{+3.18}.
These superior performances are achieved with \textbf{25\% fewer labeled samples}, highlighting CABIN's exceptional sample efficiency.
Class-wise accuracy further reveals that improvements mainly concentrate on classes where the original models struggle.
Similar trends of robustness hold on the other four datasets (Table~\ref{tab:constract_experiments_3}), e.g., a \textbf{+2.44\%} OA boost for CLOLN on HongHu.
Despite some minor trade-offs, e.g., SSFTT’s AA on PaviaU, the general results confirm CABIN’s ability to elevate performance and label-efficiency of existing supervised HSI classifiers.
Results consistently demonstrate that CABIN is effective, model-agnostic and plug-and-play.

\begin{table}[ht!]
\centering
\small
\setlength{\tabcolsep}{3.8pt}{
\begin{tabular*}{\linewidth}{cc||ccc|ccc}
\toprule
\multicolumn{2}{c||}{\textbf{Module}} & \multicolumn{3}{c|}{\textbf{Indian Pines}} & \multicolumn{3}{c}{\textbf{Salinas}} \\
\midrule
\multirow{2}{*}{UGDSS} & \multirow{2}{*}{FADS} & OA & AA & $\kappa$ & OA & AA & $\kappa$ \\
& & (\%) & (\%) & $\times 100$ & (\%) & (\%) & $\times 100$\\
\midrule
\midrule
$\times$   & $\times$    & 87.35 & 93.65 & 85.48 & 96.00 & 98.27 & 95.55 \\
\checkmark & $\times$    & 89.80 & 93.46 & 88.30 & 97.21 & \textbf{98.97} & 96.89 \\
$\times$   & \checkmark & 88.87 & 93.66 & 87.27 & 96.46 & 98.58 & 96.06 \\
\rowcolor{gray!30}\checkmark & \checkmark & \textbf{90.08} & \textbf{94.20} & \textbf{88.66} & \textbf{97.51} & 98.90 & \textbf{97.23} \\
\bottomrule
\end{tabular*}
}
\caption{Module-wise ablation study of CABIN on Indian Pines and Salinas datasets under the SSFTT baseline. }
\label{tab:ablation_noscale}
\end{table}

\begin{table*}[t]
    \centering
    
    \small
    \setlength{\tabcolsep}{4.8pt}
        \begin{tabular*}{\linewidth}{c|c||ccc|c>{\columncolor{gray!30}}cc>{\columncolor{gray!30}}c|c>{\columncolor{gray!30}}cc>{\columncolor{gray!30}}c}
            \toprule
            &
            & \multicolumn{3}{c|}{}
            & \multicolumn{4}{c|}{\textbf{CNN-based}} 
            & \multicolumn{4}{c}{\textbf{Transformer-based}} \\
            \midrule
            
            Datasets
            & Metrics
            & \multicolumn{1}{c}{SVM} 
            & \multicolumn{1}{c}{2DCNN} 
            & \multicolumn{1}{c|}{3DCNN} 
            & \multicolumn{1}{c}{$\mathrm {ReS^2}$} 
            & \multicolumn{1}{>{\columncolor{gray!30}}c}{\textbf{CABIN}} 
            & \multicolumn{1}{c}{CLOLN} 
            & \multicolumn{1}{>{\columncolor{gray!30}}c|}{\textbf{CABIN}} 
            & \multicolumn{1}{c}{SSFTT} 
            & \multicolumn{1}{>{\columncolor{gray!30}}c}{\textbf{CABIN}} 
            & \multicolumn{1}{c}{GSC-ViT} 
            & \multicolumn{1}{>{\columncolor{gray!30}}c}{\textbf{CABIN}}
            \\
            \midrule
            \midrule
            \multirow{6}{*}{Salinas}         & \multirow{2}{*}{OA(\%)} & \multirow{2}{*}{85.44} & \multirow{2}{*}{86.60 } & \multirow{2}{*}{89.70} & \multirow{2}{*}{94.78} & \textbf{95.84} & \multirow{2}{*}{91.27} & \textbf{93.19 } & \multirow{2}{*}{96.00} & \textbf{97.51} & \multirow{2}{*}{95.05} & \textbf{96.52} \\
                                &    &       &       &       &       & \textcolor[rgb]{0.0, 0.6, 0.0}{+1.06} &       & \textcolor[rgb]{0.0, 0.6, 0.0}{+1.92} &       & \textcolor[rgb]{0.0, 0.6, 0.0}{+1.51} &       & \textcolor[rgb]{0.0, 0.6, 0.0}{+1.47} \\
                                & \multirow{2}{*}{AA(\%)} & \multirow{2}{*}{90.31} & \multirow{2}{*}{92.89 } & \multirow{2}{*}{94.88} & \multirow{2}{*}{98.07} & \textbf{98.32} & \multirow{2}{*}{96.20 } & \textbf{96.50} & \multirow{2}{*}{98.27} & \textbf{98.90} & \multirow{2}{*}{98.03} & \textbf{98.21} \\
                                &    &       &       &       &       & \textcolor[rgb]{0.0, 0.6, 0.0}{+0.25} &       & \textcolor[rgb]{0.0, 0.6, 0.0}{+0.3} &       & \textcolor[rgb]{0.0, 0.6, 0.0}{+0.63} &       & \textcolor[rgb]{0.0, 0.6, 0.0}{+0.18} \\
                                &\multirow{2}{*}{$\kappa \times 100$} & \multirow{2}{*}{83.85} & \multirow{2}{*}{85.10} & \multirow{2}{*}{84.41} & \multirow{2}{*}{94.21} & \textbf{95.37} & \multirow{2}{*}{90.33} & \textbf{92.44} & \multirow{2}{*}{95.55} & \textbf{97.23} & \multirow{2}{*}{94.50} & \textbf{96.13} \\
                                &    &       &       &       &       & \textcolor[rgb]{0.0, 0.6, 0.0}{+1.16} &       & \textcolor[rgb]{0.0, 0.6, 0.0}{+2.11} &       & \textcolor[rgb]{0.0, 0.6, 0.0}{+1.68} &       & \textcolor[rgb]{0.0, 0.6, 0.0}{+1.63} \\
            \midrule
           \multirow{6}{*}{PaviaU}         & \multirow{2}{*}{OA(\%)} & \multirow{2}{*}{76.60} & \multirow{2}{*}{77.63} & \multirow{2}{*}{87.52} & \multirow{2}{*}{90.71} & \textbf{93.14} & \multirow{2}{*}{95.55} & \textbf{96.71} & \multirow{2}{*}{91.88} & \textbf{93.75} & \multirow{2}{*}{93.03} & \textbf{94.95} \\
                                &    &       &       &       &       & \textcolor[rgb]{0.0, 0.6, 0.0}{+2.43} &       & \textcolor[rgb]{0.0, 0.6, 0.0}{+1.16} &       & \textcolor[rgb]{0.0, 0.6, 0.0}{+1.87} &       & \textcolor[rgb]{0.0, 0.6, 0.0}{+1.92} \\
                                & \multirow{2}{*}{AA(\%)} & \multirow{2}{*}{83.57} & \multirow{2}{*}{81.62} & \multirow{2}{*}{90.22} & \multirow{2}{*}{92.68} & \textbf{94.08} & \multirow{2}{*}{95.45} & \textbf{96.53} & \multirow{2}{*}{93.77} & \textbf{92.85} & \multirow{2}{*}{94.05} & \textbf{95.77} \\
                                &    &       &       &       &       & \textcolor[rgb]{0.0, 0.6, 0.0}{+1.4} &       & \textcolor[rgb]{0.0, 0.6, 0.0}{+1.08} &       & \textcolor{red}{-0.92} &       & \textcolor[rgb]{0.0, 0.6, 0.0}{+1.72} \\
                                & \multirow{2}{*}{$\kappa \times 100$} & \multirow{2}{*}{70.39} & \multirow{2}{*}{71.15} & \multirow{2}{*}{84.41} & \multirow{2}{*}{87.88} & \textbf{90.99} & \multirow{2}{*}{94.12} & \textbf{95.64} & \multirow{2}{*}{89.35} & \textbf{91.70} & \multirow{2}{*}{90.76} & \textbf{93.35} \\
                                &    &       &       &       &       & \textcolor[rgb]{0.0, 0.6, 0.0}{+3.11} &       & \textcolor[rgb]{0.0, 0.6, 0.0}{+1.52} &       & \textcolor[rgb]{0.0, 0.6, 0.0}{+2.35} &       & \textcolor[rgb]{0.0, 0.6, 0.0}{+2.59} \\
            \midrule
            \multirow{6}{*}{Longkou}  & \multirow{2}{*}{OA(\%)} & \multirow{2}{*}{87.81} & \multirow{2}{*}{89.27} & \multirow{2}{*}{91.51} & \multirow{2}{*}{95.99} & \textbf{96.73} & \multirow{2}{*}{96.87} & \textbf{97.89} & \multirow{2}{*}{97.22} & \textbf{98.31} & \multirow{2}{*}{98.09} & \textbf{98.64} \\
                                & & & & & & \textcolor[rgb]{0.0, 0.6, 0.0}{+0.74} & & \textcolor[rgb]{0.0, 0.6, 0.0}{+1.02} & & \textcolor[rgb]{0.0, 0.6, 0.0}{+1.09} & & \textcolor[rgb]{0.0, 0.6, 0.0}{+0.55} \\
                                & \multirow{2}{*}{AA(\%)} & \multirow{2}{*}{84.55} & \multirow{2}{*}{86.09} & \multirow{2}{*}{93.44} & \multirow{2}{*}{96.60} & \textbf{95.55} & \multirow{2}{*}{96.90} & \textbf{97.71} & \multirow{2}{*}{96.83} & \textbf{97.23} & \multirow{2}{*}{97.43} & \textbf{97.74} \\
                                & & & & & & \textcolor{red}{-1.05} & & \textcolor[rgb]{0.0, 0.6, 0.0}{+0.81} & & \textcolor[rgb]{0.0, 0.6, 0.0}{+0.4} & & \textcolor[rgb]{0.0, 0.6, 0.0}{+0.31} \\
                                & \multirow{2}{*}{$\kappa \times 100$} & \multirow{2}{*}{84.39} & \multirow{2}{*}{86.23} & \multirow{2}{*}{90.41} & \multirow{2}{*}{94.80} & \textbf{95.74} & \multirow{2}{*}{95.93} & \textbf{97.32} & \multirow{2}{*}{96.37} & \textbf{97.78} & \multirow{2}{*}{97.49} & \textbf{98.22} \\
                                & & & & & & \textcolor[rgb]{0.0, 0.6, 0.0}{+0.94} & & \textcolor[rgb]{0.0, 0.6, 0.0}{+1.39} & & \textcolor[rgb]{0.0, 0.6, 0.0}{+1.41} & & \textcolor[rgb]{0.0, 0.6, 0.0}{+0.73} \\
                                \midrule
            \multirow{6}{*}{HongHu}  & \multirow{2}{*}{OA(\%)} & \multirow{2}{*}{77.49} & \multirow{2}{*}{79.23} & \multirow{2}{*}{81.66} & \multirow{2}{*}{87.82} & \textbf{88.32} & \multirow{2}{*}{88.47} & \textbf{90.91} & \multirow{2}{*}{87.68} & \textbf{88.75} & \multirow{2}{*}{91.52} & \textbf{92.01} \\
                                & & & & & & \textcolor[rgb]{0.0, 0.6, 0.0}{+0.5} & & \textcolor[rgb]{0.0, 0.6, 0.0}{+2.44} & & \textcolor[rgb]{0.0, 0.6, 0.0}{+1.07} & & \textcolor[rgb]{0.0, 0.6, 0.0}{+0.49} \\
                                & \multirow{2}{*}{AA(\%)} & \multirow{2}{*}{76.87} & \multirow{2}{*}{76.10} & \multirow{2}{*}{78.32} & \multirow{2}{*}{87.34} & \textbf{88.72} & \multirow{2}{*}{87.15} & \textbf{89.98} & \multirow{2}{*}{85.56} & \textbf{89.36} & \multirow{2}{*}{89.96} & \textbf{89.14} \\
                                & & & & & & \textcolor[rgb]{0.0, 0.6, 0.0}{+1.38} & & \textcolor[rgb]{0.0, 0.6, 0.0}{+2.83} & & \textcolor[rgb]{0.0, 0.6, 0.0}{+3.8} & & \textcolor{red}{-0.82} \\
                                & \multirow{2}{*}{$\kappa \times 100$} & \multirow{2}{*}{72.73} & \multirow{2}{*}{74.55} & \multirow{2}{*}{77.42} & \multirow{2}{*}{84.82} & \textbf{86.76} & \multirow{2}{*}{85.03} & \textbf{86.45} & \multirow{2}{*}{84.54} & \textbf{85.95} & \multirow{2}{*}{89.36} & \textbf{89.89} \\
                                & & & & & & \textcolor[rgb]{0.0, 0.6, 0.0}{+1.94} & & \textcolor[rgb]{0.0, 0.6, 0.0}{+1.42} & & \textcolor[rgb]{0.0, 0.6, 0.0}{+1.41} & & \textcolor[rgb]{0.0, 0.6, 0.0}{+0.53} \\

            \bottomrule

        \end{tabular*}

        \caption{
            Comparison of different types of methods on the Pavia University, WHU-Hi-Longkou, and WHU-Hi-HongHu datasets.}
    \label{tab:constract_experiments_3}
\end{table*}

\begin{figure*}[t]
    \centering
        \includegraphics[width=0.96\linewidth]{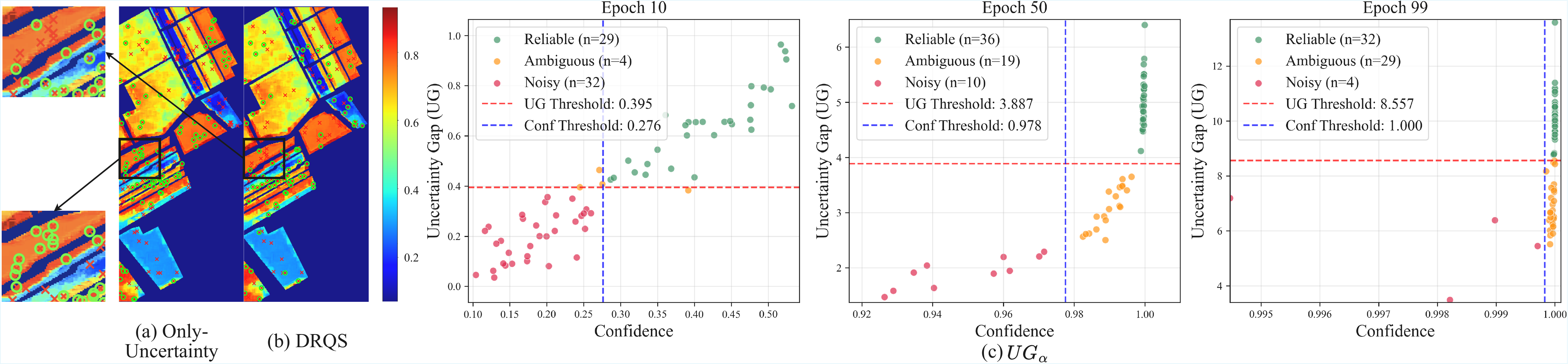}
        \caption{Left: Comparison of sampling strategies. (a) uses only dynamic thresholding, whereas (b) applies DRQS for diverse and non-redundant selection in feature space. Green circles: selected samples; red crosses: initial labeled samples. Right: Evolution of UG’s ability to discriminate sample difficulty throughout training.}
        \label{fig:sampling1}
\end{figure*}

\subsection{Ablation Study}

In this section, we conduct additional experiments to gain a deeper insights into how CABIN guides model learning more accurately and efficiently.

\subsubsection{Evaluation of Overall Effectiveness.} 
To evaluate the contribution of each module in CABIN, we perform an ablation study by selectively enabling UDGSS and FDAS across two datasets. 
As shown in Table~\ref{tab:ablation_noscale}, UGDSS alone boosts the OA from $87.35\%$ to $\textbf{89.80\%}$, while FDAS alone raises it to $\textbf{88.87\%}$ on Indian Pines, which confirms that adding either module individually can lead to notable gains.
Moreover, when we integrate UGDSS and FDAS, CABIN can achieves an OA of \textbf{90.08\%} and a $\kappa$ of \textbf{88.66} on Indian Pines, outperforming all other variants. The similar trend of robustness holds on Salinas.
These results demonstrate that UGDSS and FDAS are not only effective individually, but also complementary components that can be combined to fully realize the potential of our framework.

\subsubsection{Diversification Capability of DRQS.}
To compare the differences between the sampling strategies, we visualize the corresponding sampling results and overlay them on the uncertainty map generated by the model.
The random sampling strategy (marked by red crosses) may over-select regions where the model has learned well.
In contrast, the purely uncertainty-based method (marked by green circles) in Fig.~\ref{fig:sampling1}a) shifts the focus to the critical high-uncertainty boundary regions.
However, due to the strong spatial correlation of the HSI data, this strategy is prone to form dense and redundant sampling clusters in local regions (as shown in the enlarged illustration), resulting in diminishing learning returns.
In Fig.~\ref{fig:sampling1}b, the samples selected by DRQS (marked by green circles) are more spatially dispersed while still covering key uncertainty regions, thus capturing richer and more diverse information.
This phenomenon shows that DRQS effectively reduces spatial redundancy on the basis of retaining the discriminant regions, thereby achieving a more diverse sample selection capability.

\subsubsection{Efficiency Enhancement under Annotation Constraints.}
To evaluate the annotation efficiency of our selection strategy, we conduct an ablation study under different annotation ratios.
For the Indian Pines and Salinas datasets, the baseline method randomly selects 20 samples from each class (320 in total) for training.
Our method first uses 10 samples from each class (160 in total) for EDL pre-training, and the remaining 160 samples constitute the selection pool (i.e., the row with 100\% ratio in the Table~\ref{tab:budgets}).
According to the results, our method achieves peak performance with only 50\% of the annotations.
This demonstrates the high annotation efficiency of our strategy.
By focusing on informative and diverse samples, the strategy can capture sufficient training signals without exhaustive annotations.
When we increase the ratio to 75\% or 100\%, the performance degrades, suggesting that excessive annotation may introduce redundant or noisy supervision.
These findings confirm the effectiveness of our strategy in real-world scenarios with limited annotation resources.

\begin{table}[ht!]
\centering
\small
\setlength{\tabcolsep}{6pt}{
\begin{tabular*}{\linewidth}{c||ccc||ccc}
\toprule
 & \multicolumn{3}{c||}{\textbf{Indian Pines}} & \multicolumn{3}{c}{\textbf{Salinas}} \\
\midrule 
\midrule
\multirow{2}{*}{\textbf{Ratio}}& OA & AA & $\kappa$ & OA & AA & $\kappa$ \\
& (\%) & (\%) & $\times 100$ & (\%) & (\%) & $\times 100$\\
\midrule
0\% & 78.92 & 89.08 & 75.74 & 94.76 & 98.00 & 95.28\\
25\% & 85.81 & 91.85 & 83.78 & 96.62 & 98.72 & 96.24 \\
\rowcolor{gray!30}50\% & \textbf{90.08} & \textbf{94.20} & \textbf{88.66} & \textbf{97.51} & \textbf{98.90} & \textbf{97.23} \\
75\% & 89.01 & 94.47 & 88.43 & 96.85 & 98.73 & 96.50 \\
100\% & 89.15 & 93.58 & 87.58 & 97.11 & 98.89 & 96.79 \\
\midrule
\midrule
\multirow{2}{*}{\textbf{Number}}& OA & AA & $\kappa$ & OA & AA & $\kappa$ \\
& (\%) & (\%) & $\times 100$ & (\%) & (\%) & $\times 100$\\
\midrule
1 & 89.00&93.24 & 87.32 & 96.22 & 98.51 & 95.80\\
2 & 89.45& 93.75& 87.91 & 96.96 & 98.73& 96.61\\
3 & 89.85 & 93.61 & 88.35 & 97.14 & 98.68 & 96.82 \\
\cellcolor{gray!30}4 & 89.81& 94.07 & 88.35& \cellcolor{gray!30}\textbf{97.51} & \cellcolor{gray!30}\textbf{98.90} & \cellcolor{gray!30}\textbf{97.23}\\
5 & \textbf{90.22} & 93.76 & \textbf{88.75} & 96.84& 98.58& 96.48\\
\cellcolor{gray!30}6 & \cellcolor{gray!30}90.08 & \cellcolor{gray!30}\textbf{94.20} & \cellcolor{gray!30}88.66 & 96.22 & 98.44 & 95.80\\
7 & 89.06 & 93.72& 87.48& 96.26& 98.31 & 95.84\\
\bottomrule
\end{tabular*}
}
\caption{Performance under different annotation ratios and augmentation numbers.}
\label{tab:budgets}
\end{table}

\subsubsection{Feature Enhancement Capability of GFP.}
\begin{figure}[ht!]
    \centering
    \includegraphics[width=0.72\linewidth]{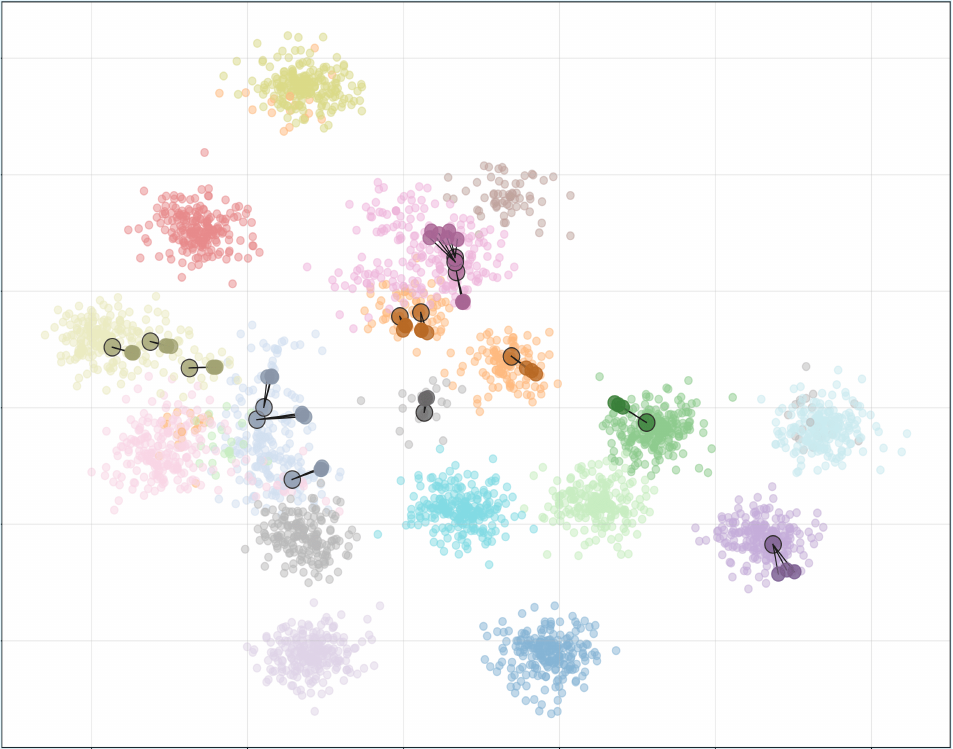}
    \caption{
    t-SNE projection of the original and augmented samples in feature space.
    Each color represents a semantic class.
    Light-colored points indicate original samples; dark points with outlines are selected samples; dark points without outlines are generated samples.
    Light and dark tones of the same color correspond to the same class.}
    \label{fig:GFP_feature}
\end{figure}

To further explore the mechanism of GFP, we conduct more experiments for qualitative and quantitative analysis.
In Figure~\ref{fig:GFP_feature}, the qualitative visualization shows the features of enhanced samples generated around the original samples.
The feature distribution of these enhanced samples does not exceed the category boundary, further strengthening the feature consistency and intra-class compactness between samples of the same label.
Quantitatively, Table~\ref{tab:budgets} shows the impact of different numbers of enhanced samples on model performance.
According to the results, although OA and $\kappa$ perform better when the number is set to 5 on the Indian Pines dataset, the overall balance is better when the number is set to 6. On the Salinas dataset, the performance peaks when the number is set to 4.
These findings confirm that moderate feature perturbation helps improve the robustness of the model, while too much may introduce noise, resulting in ambiguous decision boundary and decreased performance.
By these experiments, the effectiveness of the GFP module and the importance of controlling the intensity of feature space perturbation on the enhancement effect can be verified.

\subsubsection{Discriminative Power of $UG_\alpha$ across Training Epochs.} We track the evolution of $UG_\alpha$'s discriminative capacity over training in Fig.~\ref{fig:sampling1}c.
Early on, both softmax confidence and $UG_\alpha$ fail to separate samples by difficulty due to immature representations.
As training proceeds, confidence values saturate near 1.0, making them unreliable for identifying ambiguous or noisy cases.
In contrast, $UG_\alpha$ gradually sharpens its separation: by Epoch 50, easy samples start to emerge, and by Epoch 99, a clear boundary forms between high-UG (reliable) and low-UG (ambiguous/noisy) instances even under uniformly high confidence.
This dynamic reveals UG’s growing ability to structure the sample space meaningfully, enabling reliable group-aware training.

\section{Conclusion}
In this paper, we revisit the performance bottleneck in semi-supervised hyperspectral classification, tracing it to confirmation bias induced by over-reliance on confidence and neglect of uncertainty.
From a cognitive-behavioral consistency perspective, we interpret this as a mismatch between model perception and decision-making.
To address it, we introduce CABIN, a model-agnostic framework that closes the loop of perception, action, and correction through uncertainty-guided learning.
Experiments show that CABIN consistently improves generalization and label efficiency across diverse baselines.

\section*{Acknowledgements}
This work was supported by the National Natural Science Foundation of China (No. T2322012, No. 62572240).

\bibliography{main}

\clearpage

\appendix \input{appendix}
\end{document}

%% file: appendix.tex

\pdfinfo{
/TemplateVersion (2026.1)
}

\setcounter{secnumdepth}{0} 

\ifdefined\aaaianonymous
    \title{\textbf{Appendix}}
\else
    \title{AAAI 2026 Supplementary Material\\Camera Ready}
\fi

\newcolumntype{C}[1]{>{\centering\arraybackslash}p{#1}}
\date{} 
\maketitle

\section{Details of Datasets}
In this section, we introduce the detailed informations of the datasets used in our experiments.
These datasets are available on the relevant websites.
\begin{enumerate}
    \item Indian Pines. The Indian Pines scene, acquired by the AVIRIS sensor over northwestern Indiana, consists of $145 \times 145$ pixels with 224 spectral bands ranging from 0.4 to 2.5~$\mu$m. Following standard preprocessing, 24 water absorption bands (104--108, 150--163, and 220) were removed, resulting in 200 usable bands. The ground truth comprises 16 land-cover categories with partial class overlaps and imbalanced class distributions.

    \item Salinas. The Salinas scene, captured by the AVIRIS sensor over Salinas Valley, California, contains $512 \times 217$ pixels with a spatial resolution of 3.7~m and 224 spectral bands ranging from 400 to 2500~nm. After removing 20 water absorption bands, 204 bands were retained. The dataset includes 16 land-cover classes with 54{,}129 labeled pixels and rich spatial-spectral diversity.
    
    \item Pavia University. Collected in 2001 over the University of Pavia in northern Italy, the dataset was generated using the Reflective Optics System Imaging Spectrometer. It features 103 spectral bands and a spatial extent of 610 × 340 pixels, annotated with 9 major semantic classes.

    \item WHU-Hi-LongKou. The WHU-Hi-LongKou dataset was collected in Longkou Town, Hubei, China, using a Headwall Nano-Hyperspec sensor (8~mm focal length) mounted on a DJI Matrice 600 Pro UAV. The hyperspectral image contains $550 \times 400$ pixels with 270 bands spanning 400–1000~nm and a spatial resolution of approximately 0.463~m. The dataset includes 9 land-cover classes with 204{,}542 labeled pixels.
    
    \item WHU-Hi-HongHu. The HongHu dataset was acquired by a UAV in HongHu City, China, in 2017. It contains $940 \times 475$ pixels with 270 spectral bands covering 400-000~nm and a spatial resolution of approximately 0.04~m. The imagery is annotated with 22 representative crop classes, which exhibit high intra-class variability.

\end{enumerate}

\section{Experiment Setting}
This section details the dataset partitioning strategy, hardware and software environment used in our experiments, as well as the detailed hyperparameter settings for each model on different datasets to ensure fair comparison.

\subsection{Dataset Partitioning Strategy}

For all datasets, to ensure that all methods are compared under fair conditions, we use a fixed-sample random sampling strategy to partition the datasets into training, validation, and test sets.
Specifically, 20 samples from each class are used for training, 20 for validation, and the remaining samples are used as the test set.
This few-shot setting is designed to simulate real-world scenarios where annotation is expensive and highlights the advantage of our approach in efficiently utilizing data.
For the Indian Pines dataset, since the number of class samples is extremely uneven and there are cases where the number of classes is less than 20, we perform special processing on these samples: retaining 5 samples for the test set, 2 for validation, and the remaining samples for training.

In addition, we apply PCA-based spectral dimensionality reduction as a unified preprocessing step, retaining the top 30 principal components.
\begin{table}[h!]
    \centering
    \small
    \setlength{\tabcolsep}{1.1pt}
    \begin{tabular}{c|c|ccc|c}
        \toprule
        \multirow{2}{*}{Dataset} & \multirow{2}{*}{$w/o$ CABIN} &  \multicolumn{3}{c|}{$w/$CABIN}  & \multirow{2}{*}{Test}  \\
        & & Pretrain & Sampling & Retrain & \\
        \midrule
        \midrule
        Indian Pines & 313 & 160 & 80 & 240 & 9651 \\
        Salinas & 320 & 160 & 80 & 240 & 53489 \\
        Pavia University & 180 & 90 & 45 & 145 & 42416 \\
        LongKou & 180 & 90 & 45 & 145 & 204182  \\
        HongHu & 440 & 220 & 110 & 330 & 385813 \\
        \bottomrule
    \end{tabular}
    \caption{Details of the dataset partitioning used in the comparative experiments.}
    \label{tab:data_information}
\end{table}

\subsection{Experimental Environment}

All our experiments are performed on a workstation equipped with a single AMD EPYC 7K62 CPU and an NVIDIA A100 GPU.
Experiments are conducted under Ubuntu 24.04 with CUDA 12.6 and PyTorch 2.7.1.

\begin{table*}[t!]
    \centering
    \small
    \begin{tabular}{C{2cm}||C{4cm}||C{1cm}C{1cm}C{1cm}C{1cm}C{1cm}}
        \toprule
        \textbf{Architecture} & \textbf{Settings} & \textbf{IP}  & \textbf{LK} & \textbf{HH} & \textbf{PU} & \textbf{SA} \\
        \midrule
        \midrule
        \multirow{11}{*}{\textbf{SSFTT}} & \multicolumn{6}{c}{\textbf{Vanilla}} \\
        \cmidrule(lr){2-7}
        & Learning Rate & \multicolumn{5}{c}{$1e-4$} \\
        & Weight Decay & \multicolumn{5}{c}{0} \\
        & Patch Size & \multicolumn{4}{c}{$13\times13$} & $15\times15$ \\
        & Training Time (s) & 17.49 & 13.51 & 22.66 & 13.75 & 14.83\\
        \cmidrule(lr){2-7}
        & \multicolumn{6}{c}{\textbf{$with$ CABIN}}  \\
        \cmidrule(lr){2-7}
        & Weight Decay (\textit{Pretraining}) & \multicolumn{5}{c}{0} \\
        & Weight Decay (\textit{Retraining}) & \multicolumn{5}{c}{$5e-3$} \\
        & Aug. Samples & 6 & 5 & 6 & 4 & 4 \\
        & Training Time (s) & 24.21 & 34.08 & 36.69 & 20.51 & 25.45\\
        \midrule
        \midrule
        \multirow{11}{*}{\textbf{GSC-ViT}} & \multicolumn{6}{c}{\textbf{Vanilla}} \\
        \cmidrule(lr){2-7}
        & Learning Rate & \multicolumn{5}{c}{$1e-3$} \\
        & Weight Decay & \multicolumn{5}{c}{0} \\
        & Patch Size & \multicolumn{3}{c}{$10\times10$} & $8\times8$ & $12\times12$ \\
        & Training Time (s) & 20.59 & 16.24 & 44.13 & 19.24 & 46.47 \\
        \cmidrule(lr){2-7}
        & \multicolumn{6}{c}{\textbf{$with$ CABIN}}  \\
        \cmidrule(lr){2-7}
        & Weight Decay (\textit{Pretraining})  & \multicolumn{5}{c}{0} \\
        & Weight Decay (\textit{Retraining})  & \multicolumn{5}{c}{$5e-3$} \\
        & Aug. Samples & 5 & 4 & 4 & 3 & 5 \\
        & Training Time (s) & 27.30 & 20.15 & 56.52 & 27.61 & 59.18\\
        \bottomrule
    \end{tabular}
    \caption{
        Training configurations and runtime comparisons for \textbf{SSFTT} and \textbf{GSC-ViT} backbones under Vanilla and CABIN-enhanced settings across five hyperspectral datasets.}
    \label{tab:SSFTT}
\end{table*}

\subsection{Setup for Model-Agnostic Validation}

To verify the model-agnostic and general applicability of the proposed \textbf{CABIN} framework, we conduct comparative experiments on four different backbone networks.
For each backbone, we evaluate the following two training methods:

\textbf{Baseline (w/o CABIN)}: The model is trained via a standard supervised learning pipeline on the full labeled training set (as shown in Table~\ref{tab:data_information}). 
No uncertainty estimation or sample selection is involved.

\textbf{Ours (w/ CABIN)}: The model is trained using our proposed three-stage framework, which includes:

\begin{itemize}
    \item \textbf{Stage 1. Pretraining}: A small number of labeled samples (10 per class; the exact count in Table~\ref{tab:data_information}) are randomly selected from the training set to train an initial model equipped with uncertainty estimation capability.
    
    \item \textbf{Stage 2. Uncertainty-Guided Sampling}:
    The pretrained model is employed to estimate the uncertainty of the remaining unlabeled samples.
    Following our UGDSS strategy, we extract two disjoint subsets: a high-uncertainty subset $D_{\mathrm{hu}}$ selected for manual annotation, and a low-uncertainty, reliable subset $D_{\mathrm{re}}$ assigned with pseudo labels.
    The number of labeled samples is shown in Table~\ref{tab:data_information}.

    \item \textbf{Stage 3. Retraining}: The labeled set, comprising the initial samples and the annotated $D_{\mathrm{hu}}$, forms a compact, high-quality training set (total size detailed in Table~\ref{tab:data_information}).
    This set, along with the pseudo-labeled subset $D_{\mathrm{re}}$, is then used to retrain the model under the FDAS strategy.
\end{itemize}

For fairness, we follow the original architecture design of each backbone and tune key hyperparameters (e.g., learning rate, weight decay, and CABIN-specific parameters) per setting, while keeping shared configurations fixed: AdamW optimizer, batch size of 48, and 100 training epochs.

\begin{table}[h!]
    \centering
    \small
    \setlength{\tabcolsep}{6pt}
    \renewcommand{\arraystretch}{1.2}
    \begin{tabular}{ccccccc}
        \toprule
        & \multicolumn{1}{c}{\textbf{Parameter}} & \textbf{IP} & \textbf{SA} & \textbf{PU} & \textbf{LK} & \textbf{HH} \\
        \midrule
        \midrule
        \multirow{2}{*}{\textbf{Share}} & Learning Rate & \multicolumn{5}{c}{5e-3} \\
        \cline {2-7}
        & Weight Decay  & \multicolumn{5}{c}{0} \\
        \midrule
        \textbf{Patch Size} & — & \multicolumn{5}{c}{$11\times11$} \\
        \midrule
        \textbf{Sampling} & Aug Number & 4 & 3 & 4 & 5 & 5 \\
        \bottomrule
    \end{tabular}
    \caption{Parameter settings for the \textbf{Res\_LS$^2$CM} backbone.}
    \label{tab:Res}
\end{table}

\subsubsection{Res\_LS$^2$CM.}
A lightweight residual network that uses a lightweight spectral-spatial convolution module to build the basic residual block, with a fixed input patch spatial size of 11 × 11.
Other parameters are showen in Table~\ref{tab:Res}.

\subsubsection{CLOLN.}
The model is a channel-layer-oriented lightweight network designed to alleviate the number of parameters and computational complexity associated with CNNs. 
Other parameters are showen in Table~\ref{tab:CLOLN}.

\begin{table}[h!]
    \centering
    \small
    \setlength{\tabcolsep}{1pt}
    \renewcommand{\arraystretch}{1.2}
    \begin{tabular}{ccccccc}
        \toprule
        & \multicolumn{1}{c}{\textbf{Parameter}} & \textbf{IP} & \textbf{SA} & \textbf{PU} & \textbf{LK} & \textbf{HH} \\
        \midrule
        \midrule
        \multirow{4}{*}{\textbf{Share}} & Learning & \multicolumn{5}{c}{\multirow{2}{*}{1e-3}}\\
        & Rate & & & & & \\
        \cline {2-7}
        & Weight  & \multicolumn{5}{c}{\multirow{2}{*}{0}} \\
        & Decay & & & & & \\
        \midrule
        \textbf{Patch Size} & — & $13\times13$ & $13\times13$ & $13\times13$ & $11\times11$ & $11\times11$ \\
        \midrule
        \multirow{2}{*}{\textbf{Sampling}} & Aug & \multirow{2}{*}{4} & \multirow{2}{*}{4} & \multirow{2}{*}{4} & \multirow{2}{*}{5} & \multirow{2}{*}{5} \\
        & Number & & & & & \\
        \bottomrule
    \end{tabular}
    \caption{Parameter settings for the \textbf{CLOLN} backbone.}
    \label{tab:CLOLN}
\end{table}

\subsubsection{GSC-ViT.}
The model adopts a groupwise separable convolution ViT to capture local and global spectral-spatial information for HSI classification.
Other parameters are showen in Table~\ref{tab:SSFTT}.

\begin{figure*}[t]
    \centering
    \includegraphics[width=\linewidth]{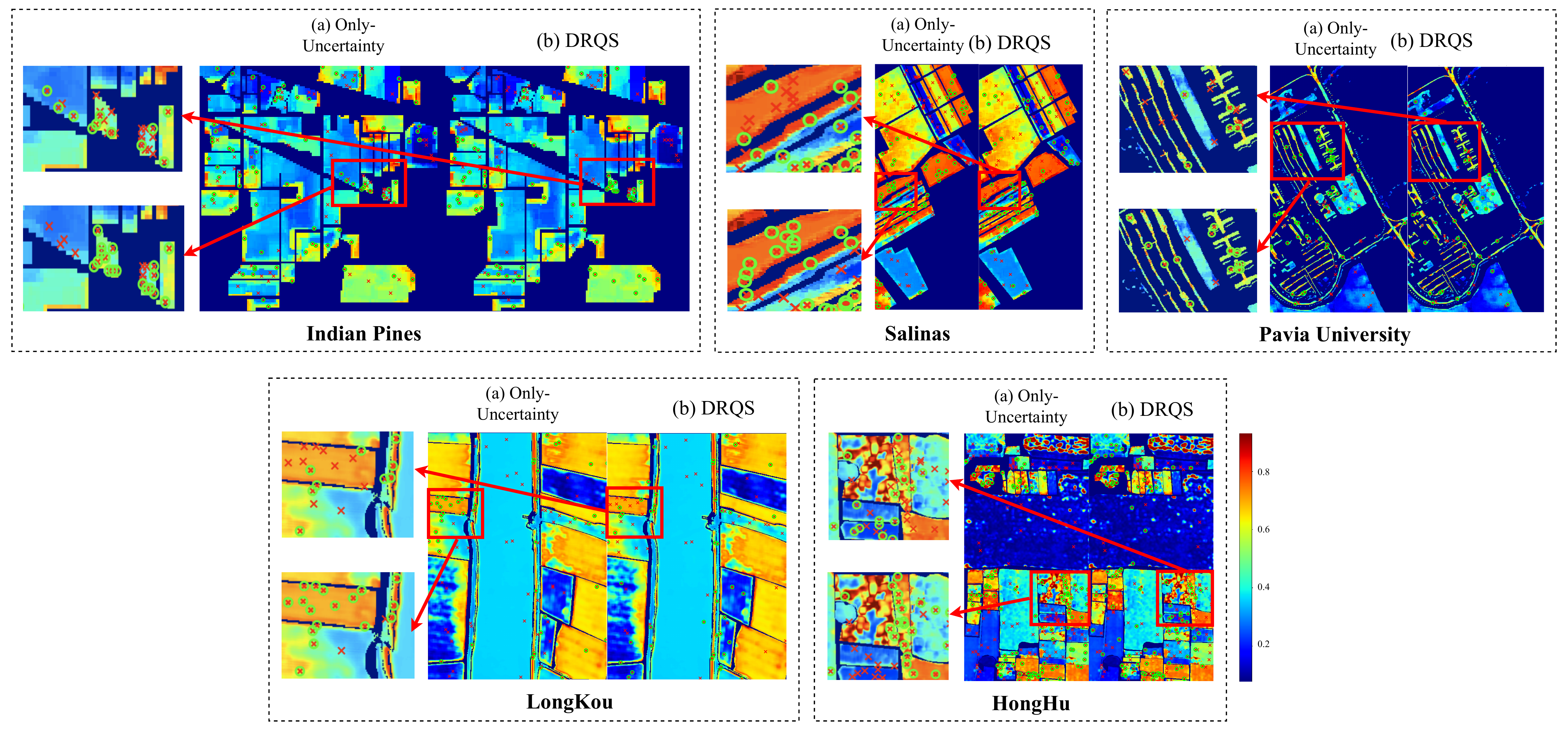}
    \caption{
        Cross-dataset comparison of sampling strategies. 
        (a) employs only dynamic thresholding, whereas (b) incorporates DRQS to promote diversity and reduce redundancy in the feature space.}
    \label{fig:SSFTT_Sampling}
\end{figure*}

\begin{figure*}[!htbp]
    \centering
    \includegraphics[width=1\linewidth]{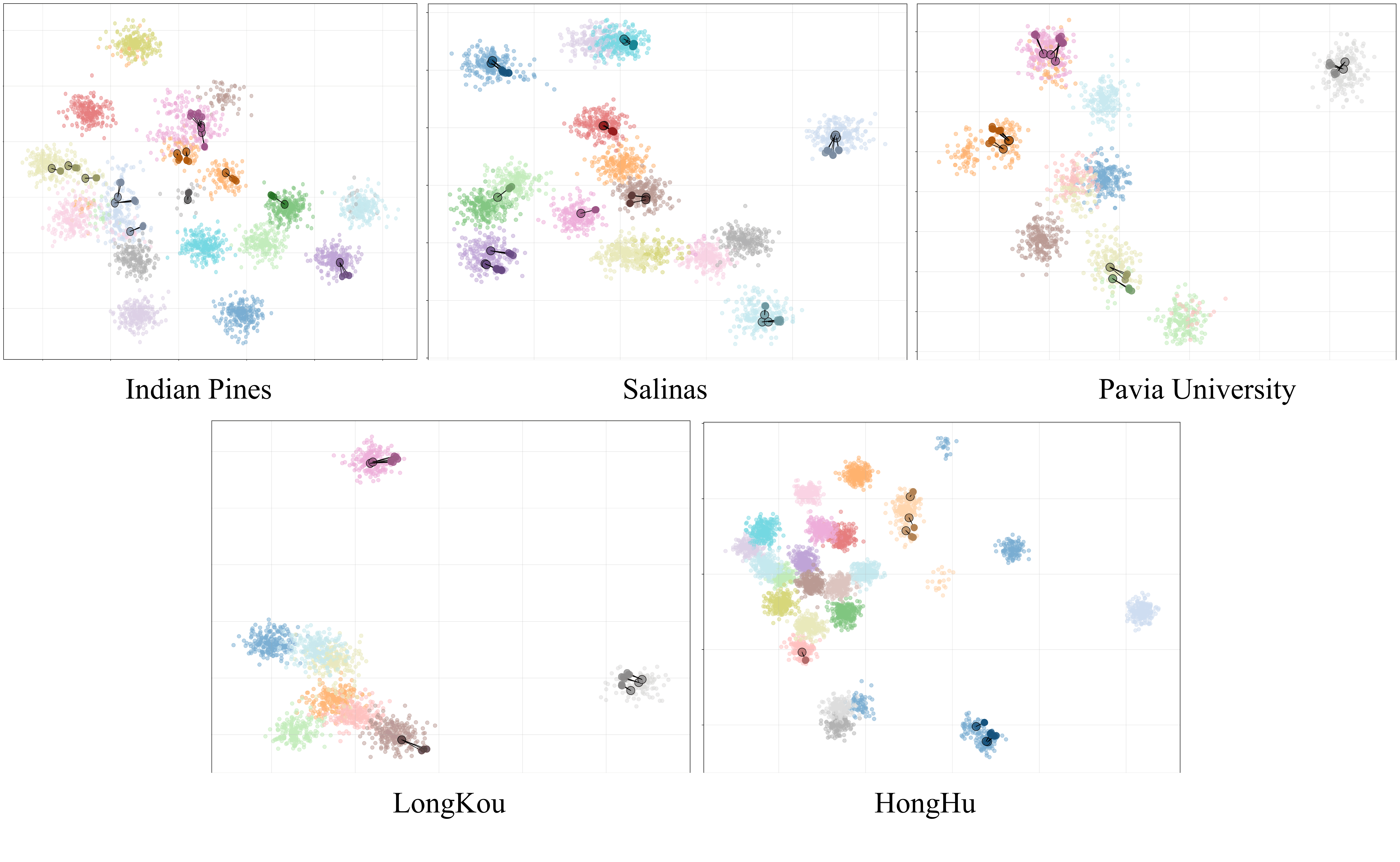}
    \caption{
        t-SNE projection of the original and augmented samples in feature space across multiple datasets. 
        Each color represents a semantic class.
        Light-colored points indicate original samples; dark points with outlines are selected samples; dark points without outlines are generated samples.
        Light and dark tones of the same color correspond to the same class.}
    \label{fig:SSFTT_tsne}
\end{figure*}

\begin{figure*}[!htbp]
    \centering
    \includegraphics[width=1\linewidth]{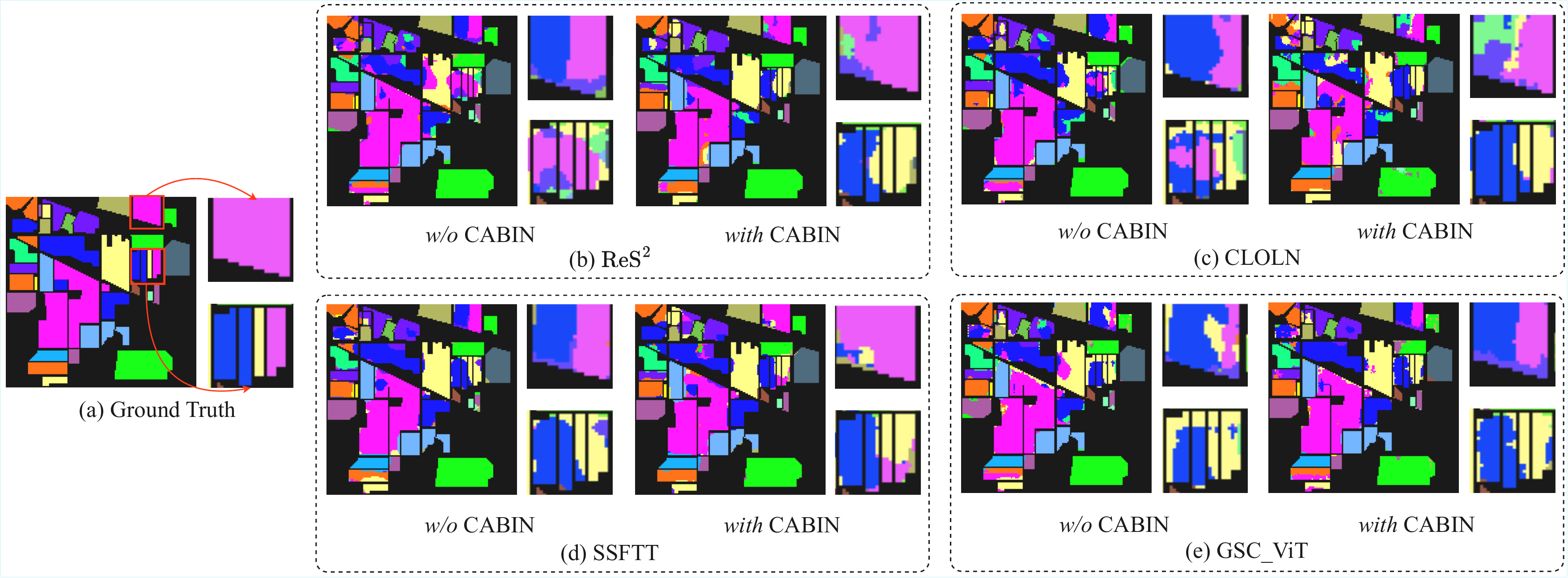}
    \caption{Comparisons of the classification maps of different methods on the \textit{Indian Pines} dataset.}
    \label{fig:IP_fig}
\end{figure*}

\subsubsection{SSFTT.}
The model consists of a 3-D convolutional layer (8 kernels, $3\times3\times3$), a 2-D convolutional layer (64 kernels, $3\times3$), a transformer encoder with 4 tokens and 4 attention heads, followed by a dropout and a linear classifier.
Other parameters are showen in Table~\ref{tab:SSFTT}.


\section{Analysis of Comparative Experimental Results}

This section provides more detailed experimental data and visualizations to complement the analysis in the main text.
Specifically, We conduct an in-depth analysis from two perspectives:
(1) quantitative classification performance, including overall metrics and accuracy per category;
and (2) qualitative classification plots, which visually demonstrate the improvement of our method in difficult-to-learn regions.

\subsubsection{Quantitative Performance Evaluation.}
Based on the quantitative experimental results showen in Table~\ref{tab:SA}-\ref{tab:HH}, we can draw the following comprehensive conclusions:
the proposed CABIN framework demonstrates consistent performance improvements across different backbone networks and five datasets, fully demonstrating its model-independence and broad applicability. 
Furthermore, as shown in Table~\ref{tab:SA}, the primary performance improvement is concentrated in difficult categories (such as class 8), where the baseline model previously performed poorly.
This demonstrates the effectiveness of our UGDSS strategy-by actively annotating high-uncertainty samples, it guides the model to focus learning resources on difficult areas.
Furthermore, for easy-to-learn categories such as class 7 or 16, CABIN maintains its original high accuracy without significant degradation.
This is due to the FDAS module's assisted learning with high-confidence pseudo-labels and sample filtering mechanism, which prevents catastrophic forgetting.
In summary, CABIN achieves targeted reinforcement of difficult categories while maintaining stable and improved overall performance, demonstrating strong robustness and generalization capabilities.

\subsubsection{Qualitative Visualization Analysis.}
\begin{figure*}[t]
    \centering
    \includegraphics[width=0.9\linewidth]{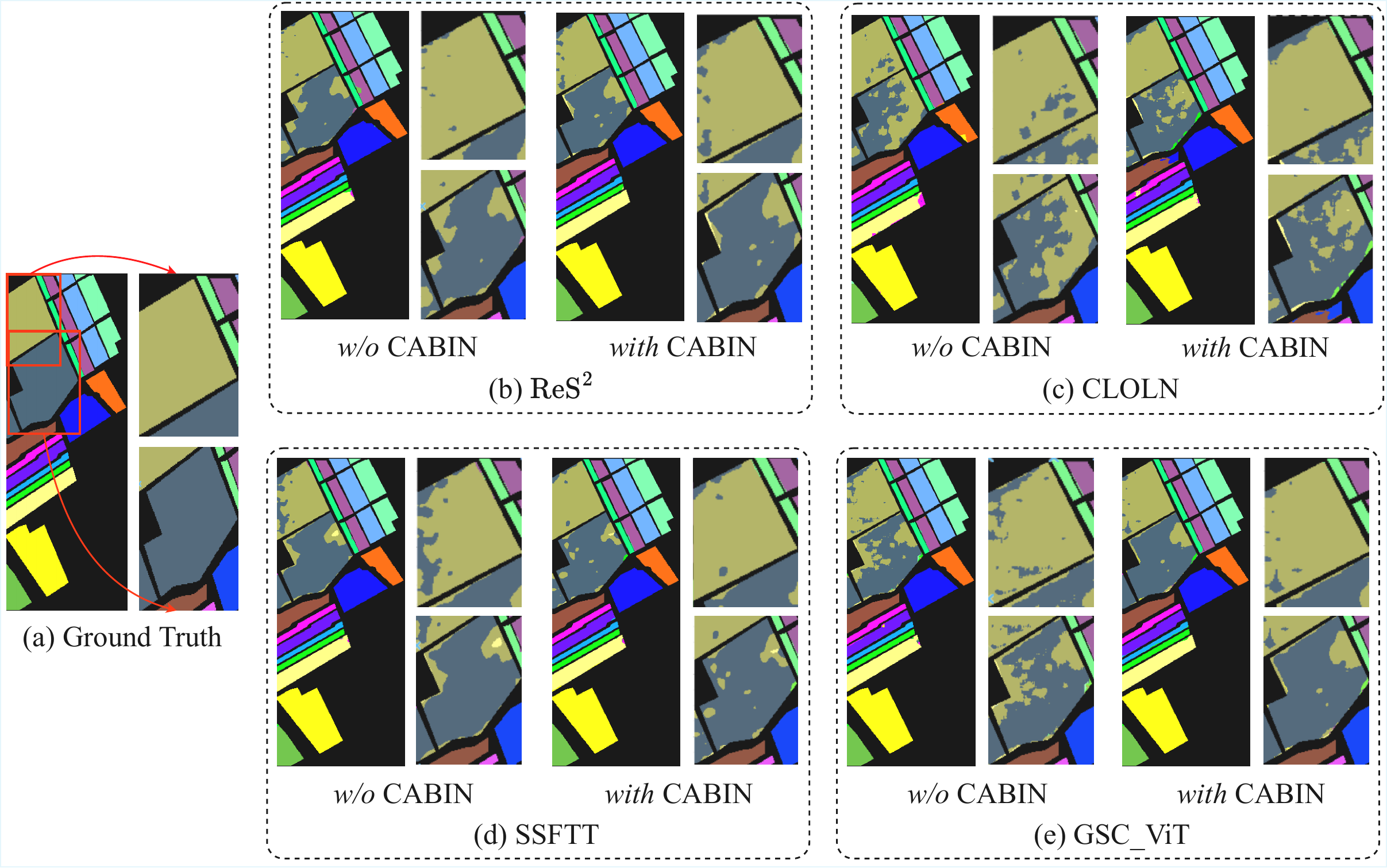}
    \caption{Comparisons of the classification maps of different methods on the \textit{Salinas} dataset.}
    \label{fig:SA_fig}
\end{figure*}

\begin{figure*}[t!]
    \centering
    \includegraphics[width=\linewidth]{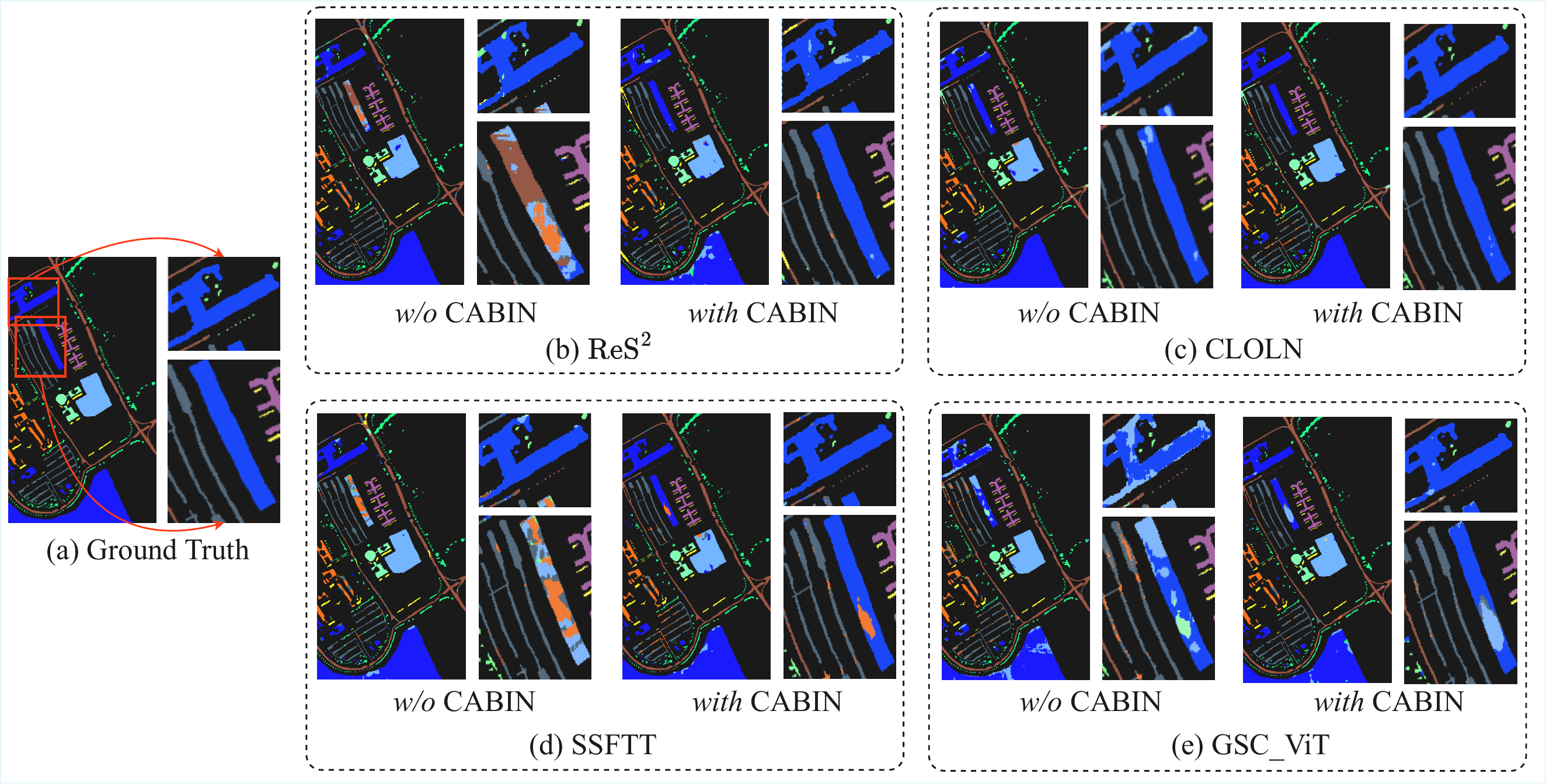}
    \caption{Comparisons of the classification maps of different methods on the \textit{Pavia University} dataset.}
    \label{fig:PU_fig}
\end{figure*}

\begin{figure*}[t!]
    \centering
    \includegraphics[width=\linewidth]{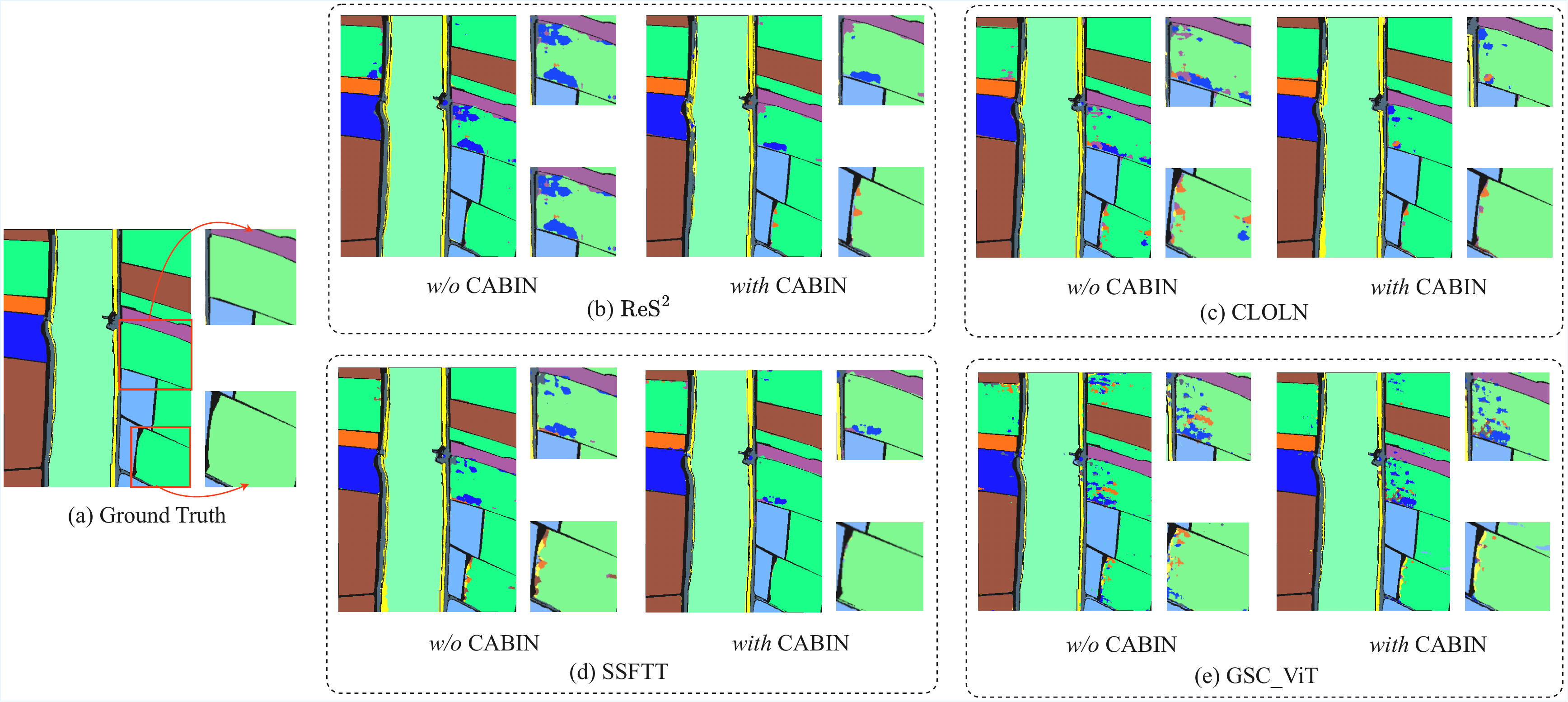}
    \caption{Comparisons of the classification maps of different methods on the \textit{LongKou} dataset.}
    \label{fig:LK_fig}
\end{figure*}

\begin{figure*}[t!]
    \centering
    \includegraphics[width=0.95\linewidth]{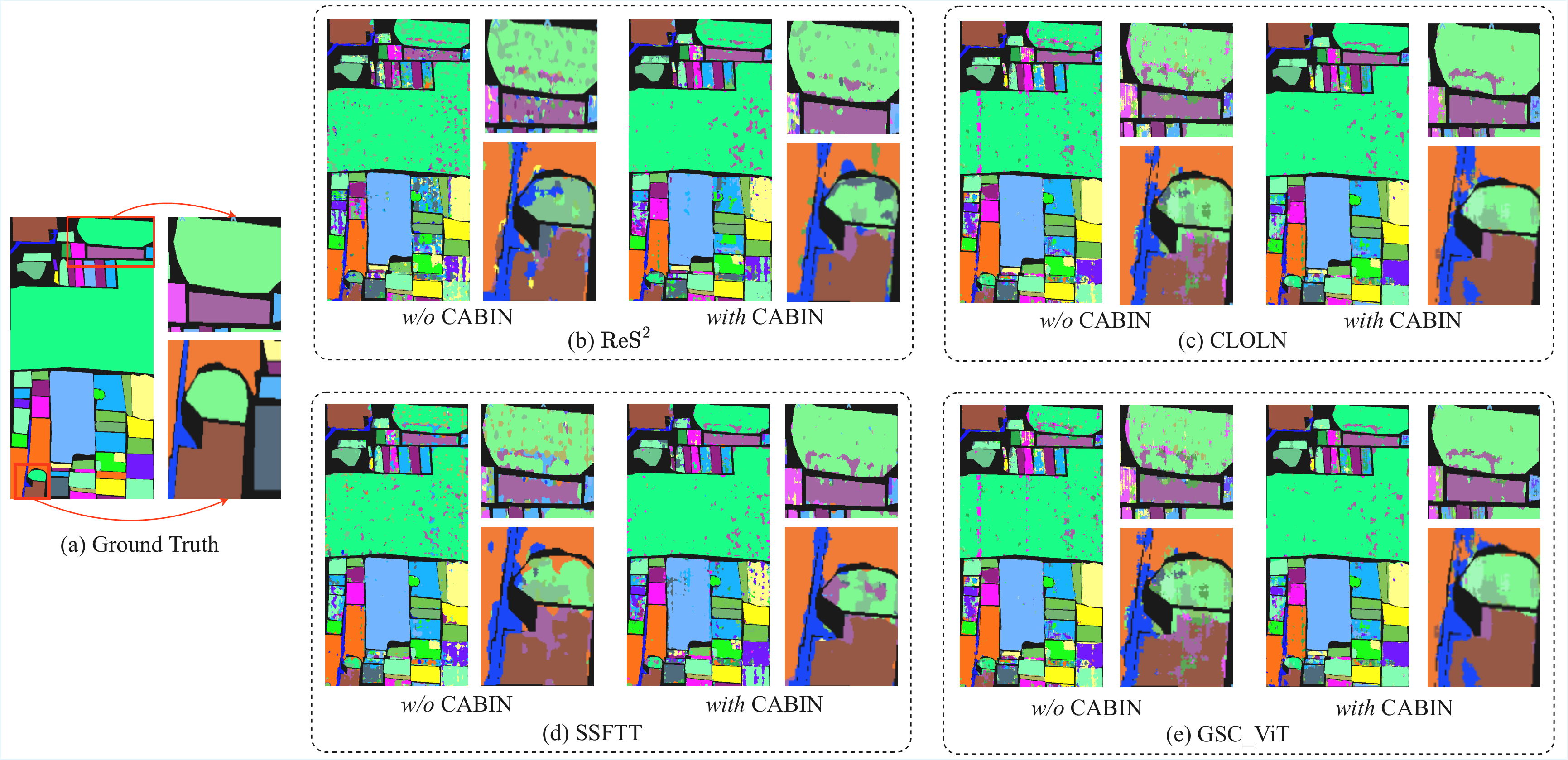}
    \caption{Comparisons of the classification maps of different methods on the \textit{HongHu} dataset.}
    \label{fig:HH_fig}
\end{figure*}

Furthermore, for a more intuitive comparison of the results, we provide visual classification maps for the five datasets in Fig.~\ref{fig:IP_fig}-\ref{fig:HH_fig}.

These qualitative results clearly demonstrate that models trained with CABIN produce more spatially consistent and accurate predictions compared to the baseline. 
Particularly, the baseline exhibits significant noise and fragmented predictions near class boundaries, indicating unstable and isolated decisions due to limited and randomly selected supervision.
In contrast, our approach produces smoother, more uniform regions that better align with the ground truth.

Overall, the detailed quantitative results and qualitative visualizations collectively validate the effectiveness of CABIN in achieving high performance by prioritizing sample value over quantity, thereby reducing annotation costs.

\section{Cross-Dataset Robustness and Generalizability Analysis of Inner Modules}
\subsection{Cross-Dataset Generalization of DRQS Sampling Diversity}

To analyze the effectiveness and diversification ability of DRQS, we conducted detailed sampling comparisons in the main experimental section. 
To further evaluate its robustness and generalizability, we extended the visualization and analysis to four additional benchmark datasets in this section.

As shown in Fig.~\ref{fig:SSFTT_Sampling}, the sampling patterns of random and uncertainty-based strategies remain largely unchanged, with random sampling tending to select samples from well-learned regions and uncertainty-based sampling focusing on uncertain boundaries. 
However, these methods often suffer from spatial redundancy.
In contrast, DRQS consistently produces more spatially dispersed and diverse selections across all datasets, effectively covering critical uncertain regions. 
This consistent pattern across multiple datasets highlights the stable performance and strong generalizability of DRQS, confirming its robustness as a sampling strategy for hyperspectral classification.

\subsection{Cross-Dataset Robustness of Gaussian Feature Perturbation}
In the main experimental section, we analyzed the feature-level behavior of GFP and its ability to preserve intra-class consistency through Gaussian perturbation.
To further assess its robustness and generalizability, we extend the qualitative visualization to four additional benchmark datasets.

As shown in Fig.~\ref{fig:SSFTT_tsne}, the enhanced samples consistently remain within their semantic class boundaries and exhibit tight clustering around the original points across all datasets.
This consistent spatial pattern indicates that GFP maintains intra-class compactness and inter-class separability regardless of the underlying data distribution.
These results collectively demonstrate that GFP is a stable and generalizable feature augmentation strategy, capable of enriching feature representations without introducing semantic drift, and thereby improving model robustness across diverse HSI domains.

\begin{table*}[t]
    \centering
    \small
    \setlength{\tabcolsep}{6.4pt}
    \begin{tabular*}{\linewidth}{c||ccc|c>{\columncolor{gray!30}}cc>{\columncolor{gray!30}}c|c>{\columncolor{gray!30}}cc>{\columncolor{gray!30}}c}
        \toprule
        & \multicolumn{3}{c|}{}
        & \multicolumn{4}{c|}{\textbf{CNN-based}}
        & \multicolumn{4}{c}{\textbf{Transformer-based}}
        \\ \midrule
        Class No.
        & \multicolumn{1}{c}{SVM} 
        & \multicolumn{1}{c}{2DCNN} 
        & \multicolumn{1}{c|}{3DCNN} 
        & \multicolumn{1}{c}{$\mathrm {ReS^2}$} 
        & \multicolumn{1}{>{\columncolor{gray!30}}c}{\textbf{CABIN}} 
        & \multicolumn{1}{c}{CLOLN} 
        & \multicolumn{1}{>{\columncolor{gray!30}}c|}{\textbf{CABIN}} 
        & \multicolumn{1}{c}{SSFTT} 
        & \multicolumn{1}{>{\columncolor{gray!30}}c}{\textbf{CABIN}} 
        & \multicolumn{1}{c}{GSC-ViT} 
        & \multicolumn{1}{>{\columncolor{gray!30}}c}{\textbf{CABIN}} \\

        \midrule
        \midrule
        1  & 92.17 &  98.05 & 99.89 & 99.95 & \textbf{100.00} & \textbf{100.00} & 87.35  & 100.00 & \textbf{100.00} & 100.00 & \textbf{100.00} \\ 
        2  & 99.25 & 96.47 & 99.84 & 100.00 & \textbf{100.00} & 83.75  & \textbf{99.95} & 100.00 & \textbf{100.00} & 100.00 & \textbf{100.00} \\ 
        3  & 79.35 & 95.19 & 93.49 & 100.00 & \textbf{100.00} & 100.00 & \textbf{100.00}  & 100.00 & \textbf{100.00} & 100.00 & \textbf{100.00} \\ 
        4  & 97.02 & 96.37 & 99.37 & 99.78 & 98.01 & \textbf{99.34} & 99.26 & 99.63 & \textbf{100.00} & 99.70 & 96.68 \\ 
        5  & 85.80 & 99.06 & 94.37 & 97.95 & \textbf{99.62} & 99.55 & \textbf{99.66} & 97.04 & \textbf{98.41} & 99.32 & \textbf{99.70} \\ 
        6  & 98.80 & 98.95 & 100.00& 99.92 & \textbf{99.97} & 100.00  & \textbf{100.00} & 100.00 & \textbf{100.00} & 100.00 & \textbf{100.00} \\ 
        7  & 96.73 & 99.54  & 99.54 & 99.97 & \textbf{100.00} & \textbf{99.58} & 99.41 & 100.00 &  \textbf{100.00} & 99.94 & \textbf{100.00} \\ 
        8  & 69.55 & 67.81 & 73.71 & 77.60 & \textbf{85.72} & 70.57 & \textbf{75.65} & 81.42 & \textbf{91.41} & 83.58 & \textbf{87.48} \\ 
        9  & 92.44 & 95.34 & 95.73 & 99.97 & \textbf{100.00} & 97.71 & \textbf{100.00} & 100.00 & \textbf{100.00} & 100.00 & \textbf{100.00} \\
        10 & 87.14 & 86.48 & 88.49 & 96.14 & \textbf{99.51} & 94.50 & \textbf{97.28} & 97.19 & \textbf{97.53} & 98.76 & \textbf{99.01} \\ 
        11 & 97.82 & 95.58 & 99.11 & 100.00 & \textbf{100.00} & \textbf{100.00}  & 92.61 & 100.00 & \textbf{100.00} & 100.00 & \textbf{100.00} \\ 
        12 & 89.14 & 98.57 & 99.84 & \textbf{99.52} & 99.47 & \textbf{99.95} & 98.83 & 96.98 & \textbf{98.89} & \textbf{99.36} & 98.57 \\
        13 & 93.13 & 97.03 & 99.75 & \textbf{100.00} & 99.32 & \textbf{100.00} &  99.89  & \textbf{100.00} & 99.09 & \textbf{100.00} & 94.29 \\
        14 & 98.10 & 94.68 & 99.34 & \textbf{100.00} & 99.13 & 99.90 & \textbf{100.00} & 99.61 & \textbf{100.00} & 97.48 & \textbf{100.00} \\
        15 & 73.23 & 63.09 & 78.85 & \textbf{99.14} & 92.50 & \textbf{94.41} & 94.05 & \textbf{97.57} & 97.04 & 90.33 & \textbf{96.04} \\
        16 & 95.29 & 97.49 & 96.77 & 99.32 & \textbf{99.94} & 100.00 & \textbf{100.00} & 98.70 & \textbf{100.00} & \textbf{100.00} & 99.55 \\
        \midrule
        \multirow{2}{*}{OA(\%)} & \multirow{2}{*}{85.44} & \multirow{2}{*}{86.60} & \multirow{2}{*}{89.70} & \multirow{2}{*}{94.78} & \textbf{95.84} & \multirow{2}{*}{91.27}  & \textbf{93.19} & \multirow{2}{*}{96.00} & \textbf{97.51} & \multirow{2}{*}{95.05} & \textbf{96.52} \\ 
               &       &       &       &       & \textcolor[rgb]{0.0, 0.6, 0.0}{+1.06} &       & \textcolor[rgb]{0.0, 0.705, 0.0}{+1.92} &       & \textcolor[rgb]{0.0, 0.6, 0.0}{+1.51} &       & \textcolor[rgb]{0.0, 0.6, 0.0}{+1.47} \\
        \multirow{2}{*}{AA(\%)} & \multirow{2}{*}{90.31} & \multirow{2}{*}{92.89} & \multirow{2}{*}{94.88} & \multirow{2}{*}{98.07} & \textbf{98.32} & \multirow{2}{*}{96.20} & \textbf{96.50}  & \multirow{2}{*}{98.27} & \textbf{98.90} & \multirow{2}{*}{98.03} & \textbf{98.21} \\
               &       &       &       &       & \textcolor[rgb]{0.0, 0.6, 0.0}{+0.25} &       & \textcolor[rgb]{0.0, 0.6, 0.0}{+0.3} &       & \textcolor[rgb]{0.0, 0.6, 0.0}{+0.63} &       & \textcolor[rgb]{0.0, 0.6, 0.0}{+0.18} \\
        \multirow{2}{*}{$\kappa \times 100$} & \multirow{2}{*}{83.85} & \multirow{2}{*}{85.10} & \multirow{2}{*}{84.41} & \multirow{2}{*}{94.21} & \textbf{95.37} & \multirow{2}{*}{90.33} & \textbf{92.44} & \multirow{2}{*}{95.55} & \textbf{97.23} & \multirow{2}{*}{94.50} & \textbf{96.13} \\
               &       &       &       &       & \textcolor[rgb]{0.0, 0.6, 0.0}{+1.16} &       & \textcolor[rgb]{0.0, 0.6, 0.0}{+2.11} &       & \textcolor[rgb]{0.0, 0.6, 0.0}{+1.68} &       & \textcolor[rgb]{0.0, 0.6, 0.0}{+1.63} \\
        \bottomrule

    \end{tabular*}

    \caption{Comparison of classification and overall performance on the \textit{
        Salinas} dataset across different methods. 
        For other methods, 20 samples per class (320 total) are used, while CABIN uses only 240 samples. 
        CABIN results are highlighted in light gray.
        Red font indicates performance degradation, while Green font indicates improvement.
        The best results for both per-class accuracy and overall metrics, comparing CABIN and non-CABIN methods, are shown in \textbf{bold}.}
    \label{tab:SA}
\end{table*}

\begin{table*}[t]
    \centering
    \small
    \setlength{\tabcolsep}{6.4pt}
    \begin{tabular*}{\linewidth}{c||ccc|c>{\columncolor{gray!30}}cc>{\columncolor{gray!30}}c|c>{\columncolor{gray!30}}cc>{\columncolor{gray!30}}c}
        \toprule
        & \multicolumn{3}{c|}{}
        & \multicolumn{4}{c|}{\textbf{CNN-based}}
        & \multicolumn{4}{c}{\textbf{Transformer-based}}
        \\ \midrule
        Class No.
        & \multicolumn{1}{c}{SVM} 
        & \multicolumn{1}{c}{2DCNN} 
        & \multicolumn{1}{c|}{3DCNN} 
        & \multicolumn{1}{c}{$\mathrm {ReS^2}$} 
        & \multicolumn{1}{>{\columncolor{gray!30}}c}{\textbf{CABIN}} 
        & \multicolumn{1}{c}{CLOLN} 
        & \multicolumn{1}{>{\columncolor{gray!30}}c|}{\textbf{CABIN}} 
        & \multicolumn{1}{c}{SSFTT} 
        & \multicolumn{1}{>{\columncolor{gray!30}}c}{\textbf{CABIN}} 
        & \multicolumn{1}{c}{GSC-ViT} 
        & \multicolumn{1}{>{\columncolor{gray!30}}c}{\textbf{CABIN}} \\
        \midrule
        \midrule
         1  & 76.24 & 70.43 & 83.21 & 86.45 & \textbf{91.76} & \textbf{90.46} & 90.23  & 93.73 & \textbf{96.02} & 92.28 & \textbf{96.19} \\ 
         2  & 71.11 & 78.97 & 85.87 & 89.45 & \textbf{93.36} & 97.13  & \textbf{99.58} & 91.97 & \textbf{96.86} & 94.75 & \textbf{93.81} \\ 
         3  & 74.43 & 66.79 & 92.74 & 83.63 & \textbf{86.35} & \textbf{86.98} & \textbf{97.23}  & \textbf{85.48} & 85.38 & \textbf{88.78} & 88.10 \\ 
         4  & 89.91 & 93.64 & 94.58 & 91.50 & \textbf{92.89} & \textbf{96.46} & 95.90 & \textbf{95.80} & 88.96 & \textbf{94.68} & 92.23 \\ 
         5  & 99.25 & 99.46 & 99.75 & \textbf{100.00} & 99.77 & 99.31 & \textbf{99.54} & 100.00 & \textbf{100.00} & 99.69 & \textbf{99.69} \\ 
         6  & 80.81 & 70.54 & 88.42 & 95.53 & \textbf{95.67} & 95.07  & \textbf{98.20} & 90.70& \textbf{92.08} & 83.66 & \textbf{99.10} \\ 
         7  & 89.31  & 84.62 & 90.49 & \textbf{100.00} & 99.88 & 99.69 & \textbf{99.92} & \textbf{100.00} &  98.22 & 99.84 & \textbf{99.10} \\ 
         8  & 71.29 & 70.53 & 78.59 & \textbf{94.43} & 89.15 & \textbf{98.76} & 89.54 & \textbf{82.84} & 79.74 & 93.00 & \textbf{94.87} \\ 
         9  & 99.87 & 99.63 & 98.37& 93.16 & \textbf{97.79} & 95.26 & \textbf{98.68} & 98.02 & \textbf{98.35} & 99.78 & \textbf{99.89} \\
         \midrule
        \multirow{2}{*}{OA(\%)} & \multirow{2}{*}{76.60} & \multirow{2}{*}{77.63} & \multirow{2}{*}{87.52} & \multirow{2}{*}{90.71} & \textbf{93.14} & \multirow{2}{*}{95.55}  & \textbf{96.71} & \multirow{2}{*}{91.88} & \textbf{93.75} & \multirow{2}{*}{93.03} & \textbf{94.95} \\ 
               &       &       &       &       & \textcolor[rgb]{0.0, 0.6, 0.0}{+2.43} &       & \textcolor[rgb]{0.0, 0.705, 0.0}{+1.16} &       & \textcolor[rgb]{0.0, 0.6, 0.0}{+1.87} &       & \textcolor[rgb]{0.0, 0.6, 0.0}{+1.92} \\
        \multirow{2}{*}{AA(\%)} & \multirow{2}{*}{83.57} & \multirow{2}{*}{81.62} & \multirow{2}{*}{90.22} & \multirow{2}{*}{92.68} & \textbf{94.08} & \multirow{2}{*}{95.45} & \textbf{96.53}  & \multirow{2}{*}{93.17} & \textbf{92.85} & \multirow{2}{*}{94.05} & \textbf{95.77} \\
               &       &       &       &       & \textcolor[rgb]{0.0, 0.6, 0.0}{+1.4} &       & \textcolor[rgb]{0.0, 0.6, 0.0}{+1.08} &       & \textcolor{red}{-0.92} &       & \textcolor[rgb]{0.0, 0.6, 0.0}{+1.72} \\
        \multirow{2}{*}{$\kappa \times 100$} & \multirow{2}{*}{70.39} & \multirow{2}{*}{71.15} & \multirow{2}{*}{84.41} & \multirow{2}{*}{87.88} & \textbf{90.99} & \multirow{2}{*}{94.12} & \textbf{95.64} & \multirow{2}{*}{89.35} & \textbf{91.70} & \multirow{2}{*}{90.76} & \textbf{93.35} \\
               &       &       &       &       & \textcolor[rgb]{0.0, 0.6, 0.0}{+3.11} &       & \textcolor[rgb]{0.0, 0.6, 0.0}{+1.52} &       & \textcolor[rgb]{0.0, 0.6, 0.0}{+2.35} &       & \textcolor[rgb]{0.0, 0.6, 0.0}{+2.59} \\
        \bottomrule
    \end{tabular*}

    \caption{Comparison of classification and overall performance on the \textit{
        PaviaU} dataset across different methods. For other methods, 20 samples per class (180 total) are used, while CABIN uses only 135 samples. CABIN results are highlighted in \colorbox{gray!30}{light gray}. \textcolor[rgb]{0.0, 0.6, 0.0}{Green font} indicates performance improvement. The best results for both per-class accuracy and overall metrics, comparing CABIN and non-CABIN methods, are shown in \textbf{bold}.}
    \label{tab:PU}
\end{table*}

\begin{table*}[t] 
    \centering
    \small
    \setlength{\tabcolsep}{6.4pt}
    \begin{tabular*}{\linewidth}{c||ccc|c>{\columncolor{gray!30}}cc>{\columncolor{gray!30}}c|c>{\columncolor{gray!30}}cc>{\columncolor{gray!30}}c}
        \toprule
        & \multicolumn{3}{c|}{}
        & \multicolumn{4}{c|}{\textbf{CNN-based}}
        & \multicolumn{4}{c}{\textbf{Transformer-based}}
        \\ \midrule
        Class No.
        & \multicolumn{1}{c}{SVM} 
        & \multicolumn{1}{c}{2DCNN} 
        & \multicolumn{1}{c|}{3DCNN} 
        & \multicolumn{1}{c}{$\mathrm {ReS^2}$} 
        & \multicolumn{1}{>{\columncolor{gray!30}}c}{\textbf{CABIN}} 
        & \multicolumn{1}{c}{CLOLN} 
        & \multicolumn{1}{>{\columncolor{gray!30}}c|}{\textbf{CABIN}} 
        & \multicolumn{1}{c}{SSFTT} 
        & \multicolumn{1}{>{\columncolor{gray!30}}c}{\textbf{CABIN}} 
        & \multicolumn{1}{c}{GSC-ViT} 
        & \multicolumn{1}{>{\columncolor{gray!30}}c}{\textbf{CABIN}} \\
        \midrule
        \midrule
         1  & 89.96 & 94.97 & 95.95 & \textbf{99.85} & 99.95 & \textbf{99.92} & 99.89  & \textbf{99.92} & 99.90 & 99.30 & \textbf{99.35} \\ 
         2  & 77.14 & 62.65 & 96.32 & \textbf{98.86} & 98.51 & 99.36  & \textbf{99.74} & 97.42 & 99.82 & 98.48 & \textbf{99.20} \\ 
         3  & 87.28 & 90.54 & 98.42 & \textbf{98.80} & 99.77 & 100.00 & \textbf{100.00} & \textbf{100.00} & 97.23 & 96.72 & \textbf{98.10} \\ 
         4  & 76.51 & 78.56 & 89.30 & 92.44 & \textbf{95.74} & 96.49 & \textbf{97.30} & 95.28 & \textbf{98.02} & 95.99 & \textbf{96.50} \\ 
         5  & 82.94 & 81.14 & 89.24 & 98.91 & \textbf{99.73} & 99.05 & \textbf{99.50} & \textbf{98.30} & 95.67 & 97.93 & \textbf{98.30} \\ 
         6  & 93.82 & 95.50 & 96.46 & 98.34 & \textbf{99.53} & \textbf{99.68}  & 99.50 & \textbf{100.00} & 99.86 & \textbf{99.26} & 99.10 \\ 
         7  & 99.90  & 99.61 & 99.02 & 96.84 & \textbf{97.27} & 95.62 & \textbf{97.90} & 97.92 & \textbf{98.44} & \textbf{99.92} & 97.95 \\ 
         8  & 81.69 & 88.78 & 89.27 & \textbf{94.85} & 81.85 & 92.80 & \textbf{94.10} & \textbf{96.06} & 94.01 & 93.01 & \textbf{94.30} \\ 
         9  & 71.86 & 83.18 & 87.04 & \textbf{90.60} & 87.67 & 89.21 & \textbf{92.80} & 86.64 & \textbf{92.16} & 96.20 & \textbf{97.90} \\
        \midrule
        \multirow{2}{*}{OA(\%)} & \multirow{2}{*}{87.81} & \multirow{2}{*}{89.27} & \multirow{2}{*}{91.52} & \multirow{2}{*}{\textbf{95.99}} & \textbf{96.73} & \multirow{2}{*}{96.87}  & \textbf{97.89} & \multirow{2}{*}{97.22} & \textbf{98.31} & \multirow{2}{*}{98.09} & \textbf{98.64} \\ 
               &       &       &       &       & \textcolor[rgb]{0.0, 0.6, 0.0}{+0.74} &       & \textcolor[rgb]{0.0, 0.705, 0.0}{+1.02} & & \textcolor[rgb]{0.0, 0.6, 0.0}{+1.09} &       & \textcolor[rgb]{0.0, 0.6, 0.0}{+0.55} \\
        \multirow{2}{*}{AA(\%)} & \multirow{2}{*}{84.55} & \multirow{2}{*}{86.09} & \multirow{2}{*}{93.44} & \multirow{2}{*}{\textbf{96.60}} & \textbf{95.55} & \multirow{2}{*}{96.90} & \textbf{97.71}  &\multirow{2}{*}{96.83} & \textbf{97.23} & \multirow{2}{*}{97.43} & \textbf{97.74} \\
               &       &       &       &       & \textcolor{red}{-1.05} &       & \textcolor[rgb]{0.0, 0.6, 0.0}{+0.81}& & \textcolor[rgb]{0.0, 0.6, 0.0}{+0.4} &       & \textcolor[rgb]{0.0, 0.6, 0.0}{+0.31} \\
        \multirow{2}{*}{$\kappa \times 100$} & \multirow{2}{*}{84.39} & \multirow{2}{*}{86.23} & \multirow{2}{*}{90.41} & \multirow{2}{*}{\textbf{94.80}} & \textbf{95.74} & \multirow{2}{*}{95.93} & \textbf{97.32} & \multirow{2}{*}{96.37} & \textbf{97.78} & \multirow{2}{*}{97.49} & \textbf{98.22} \\
               &       &       &       &       & \textcolor[rgb]{0.0, 0.6, 0.0}{+0.94} &    & \textcolor[rgb]{0.0, 0.6, 0.0}{+1.39} &  & \textcolor[rgb]{0.0, 0.6, 0.0}{+1.41} &       & \textcolor[rgb]{0.0, 0.6, 0.0}{+0.73} \\
        \bottomrule

    \end{tabular*}

    \caption{Comparison of classification and overall performance on the \textit{LongKou} dataset across different methods. For other methods, 20 samples per class (180 total) are used, while CABIN uses only 135 samples. CABIN results are highlighted in \colorbox{gray!30}{light gray}. \textcolor[rgb]{0.0, 0.6, 0.0}{Green font} indicates performance improvement. The best results for both per-class accuracy and overall metrics, comparing CABIN and non-CABIN methods, are shown in \textbf{bold}.}
    \label{tab:LK}
\end{table*}

\begin{table*}[t]      
    \centering     
    \small     
    \setlength{\tabcolsep}{6.4pt}     
    \begin{tabular*}{\linewidth}{c||ccc|c>{\columncolor{gray!30}}cc>{\columncolor{gray!30}}c|c>{\columncolor{gray!30}}cc>{\columncolor{gray!30}}c}         
        \toprule         
        & \multicolumn{3}{c|}{}         
        & \multicolumn{4}{c|}{\textbf{CNN-based}}         
        & \multicolumn{4}{c}{\textbf{Transformer-based}}  \\ 
        \midrule         
        Class No.         
        & \multicolumn{1}{c}{SVM}          
        & \multicolumn{1}{c}{2DCNN}          
        & \multicolumn{1}{c|}{3DCNN}          
        & \multicolumn{1}{c}{$\mathrm {ReS^2}$}          
        & \multicolumn{1}{>{\columncolor{gray!30}}c}{\textbf{CABIN}}         
        & \multicolumn{1}{c}{CLOLN}          
        & \multicolumn{1}{>{\columncolor{gray!30}}c|}{\textbf{CABIN}}          
        & \multicolumn{1}{c}{SSFTT}          
        & \multicolumn{1}{>{\columncolor{gray!30}}c}{\textbf{CABIN}}          
        & \multicolumn{1}{c}{GSC-ViT}          
        & \multicolumn{1}{>{\columncolor{gray!30}}c}{\textbf{CABIN}} \\         
        \midrule         
        \midrule          
        1  & 85.67 & 90.76 & 92.66 & \textbf{97.80} & 92.02 & 91.74 & \textbf{94.50}  & \textbf{96.01} & 93.62 & 94.80 & \textbf{94.82} \\           
        2  & 86.37 & 86.45 & 86.45 & 74.33 & \textbf{80.59} & \textbf{87.24} & 82.15  & \textbf{88.65} & 87.73 & 80.62 & \textbf{83.44} \\           
        3  & 77.16 & 78.52 & 75.76 & 86.25 & \textbf{89.80} & 83.67 & \textbf{89.90}  & 85.96 & \textbf{86.98} & \textbf{83.87} & 80.48 \\           
        4  & 81.40 & 86.62 & 88.56 & 96.60 & \textbf{97.14} & 93.05 & \textbf{94.35}  & \textbf{96.32} & 94.78 & \textbf{96.16} & 95.19 \\           
        5  & 78.68 & 78.53 & 78.94 & 87.12 & \textbf{88.94} & \textbf{94.63} & 91.64  & 85.32 & \textbf{92.75} & 81.85 & \textbf{90.89} \\          
        6  & 81.53 & 80.92 & 89.27 & 92.58 & \textbf{92.99} & 91.17 & \textbf{93.12} & \textbf{89.95} & 83.72 & \textbf{94.73} & 92.66 \\           
        7  & 62.30 & 54.44 & 60.03 & 73.90 & \textbf{76.68} & 75.94 & \textbf{78.50}  & 66.63 & \textbf{67.26} & 80.10 & \textbf{84.84} \\           
        8  & 41.93 & 34.99 & 38.76 & 70.22 & \textbf{75.60} & 69.66 & \textbf{81.42}  & 86.60 & \textbf{89.01} & 75.78 & \textbf{86.65} \\           
        9  & 94.08 & 97.25 & 94.14 & \textbf{98.66} & 94.42 & \textbf{95.14} & 93.72  & \textbf{93.93} & 93.52 & 92.52 & \textbf{98.14} \\          
        10  & 66.12 & 71.52 & 64.47 & \textbf{86.72} & 84.20 & \textbf{87.79} & 85.80  & \textbf{91.65} & 70.41 & \textbf{90.87} & 87.38 \\          
        11  & 57.84 & 44.16 & 63.64 & 84.34 & \textbf{89.21} & 75.88 & \textbf{79.80}  & 0.92  & \textbf{91.02} & \textbf{86.50} & 69.58 \\          
        12  & 60.24 & 57.95 & 58.87 & 79.12 & \textbf{82.63} & 58.85 & \textbf{86.02}  & 39.88 & \textbf{84.89} & 74.42 & \textbf{82.57} \\          
        13  & 62.94 & 63.82 & 63.07 & 83.40 & \textbf{87.21} & 58.85 & \textbf{88.05}  & 78.63 & \textbf{79.06} & 78.41 & \textbf{79.32} \\          
        14  & 77.46 & 82.67 & 76.17 & \textbf{96.78} & 93.79 & 91.38 & \textbf{92.76}  & \textbf{98.78} & 95.07 & \textbf{96.72} & 93.85 \\          
        15  & 91.18 & 83.93 & 92.33 & 68.33 & \textbf{83.02} & \textbf{91.54} & 86.55  & \textbf{100.00} & 99.90 & \textbf{99.27} & 98.06 \\          
        16  & 80.26 & 89.79 & 83.72 & 88.71 & \textbf{90.17} & \textbf{98.45} & 93.95  & \textbf{97.30} & 92.38 & \textbf{99.09} & 94.54 \\          
        17  & 81.76 & 91.25 & 82.53 & \textbf{91.92} & 89.62 & \textbf{95.93} & 90.85  & 96.90 & \textbf{97.58} & \textbf{99.26} & 98.60 \\          
        18  & 82.38 & 83.86 & 87.09 & 92.65 & \textbf{93.10} & \textbf{98.08} & 94.88  & \textbf{97.99} & 93.01 & \textbf{98.93} & 98.65 \\          
        19  & 84.78 & 80.37 & 84.78 & 83.02 & \textbf{84.26} & \textbf{98.91} & 88.88  & \textbf{97.99} & 92.94 & \textbf{89.92} & 82.32 \\          
        20  & 87.92 & 86.38 & 93.04 & 93.12 & \textbf{95.30} & 92.08 & 94.96  & \textbf{98.17} & 87.52 & 91.58 & \textbf{97.59} \\          
        21  & 82.32 & 84.19 & 80.05 & \textbf{98.08} & 96.59 & 96.21 & \textbf{96.80}  & 93.21 & \textbf{99.61} & \textbf{98.84} & 98.29 \\          
        22  & 86.77 & 66.17 & 88.62 & \textbf{97.02} & 94.83 & 96.52 & \textbf{97.30}  & \textbf{100.00} & 97.10 & 94.80 & \textbf{99.05} \\          
        \midrule         
        \multirow{2}{*}{OA(\%)} & \multirow{2}{*}{77.49} & \multirow{2}{*}{79.23} & \multirow{2}{*}{81.66} & \multirow{2}{*}{87.82} & \textbf{88.32} & \multirow{2}{*}{88.47} & \textbf{90.91} & \multirow{2}{*}{87.68} & \textbf{88.75} & \multirow{2}{*}{91.52} & \textbf{92.01} \\                 
        &       &       &       &       & \textcolor[rgb]{0.0, 0.6, 0.0}{+0.5} &       & \textcolor[rgb]{0.0, 0.705, 0.0}{+2.44} &       & \textcolor[rgb]{0.0, 0.6, 0.0}{+1.07} &       & \textcolor[rgb]{0.0, 0.6, 0.0}{+0.49} \\         
        \multirow{2}{*}{AA(\%)} & \multirow{2}{*}{76.87} & \multirow{2}{*}{76.10} & \multirow{2}{*}{78.32} & \multirow{2}{*}{87.34} & \textbf{88.72} & \multirow{2}{*}{87.15} & \textbf{89.98}  & \multirow{2}{*}{85.56} & \textbf{89.36} & \multirow{2}{*}{89.96} & 89.14 \\                
        &       &       &       &       & \textcolor[rgb]{0.0, 0.6, 0.0}{+1.38} &       & \textcolor[rgb]{0.0, 0.6, 0.0}{+2.83} &       & \textcolor[rgb]{0.0, 0.6, 0.0}{+3.8} &       & \textcolor{red}{-0.82} \\         
        \multirow{2}{*}{$\kappa \times 100$} & \multirow{2}{*}{72.73} & \multirow{2}{*}{74.55} & \multirow{2}{*}{77.42} & \multirow{2}{*}{84.82} & \textbf{86.76} & \multirow{2}{*}{85.03} & \textbf{86.45} & \multirow{2}{*}{84.54} & \textbf{85.95} & \multirow{2}{*}{89.36} & \textbf{89.89} \\                
        &       &       &       &       & \textcolor[rgb]{0.0, 0.6, 0.0}{+1.94} &       & \textcolor[rgb]{0.0, 0.6, 0.0}{+1.42} &       & \textcolor[rgb]{0.0, 0.6, 0.0}{+1.41} &       & \textcolor[rgb]{0.0, 0.6, 0.0}{+0.53} \\         
        \bottomrule     

    \end{tabular*}

    \caption{Comparison of classification and overall performance on the \textit{HongHu} dataset across different methods. For other methods, 20 samples per class (440 total) are used, while CABIN uses only 330 samples. CABIN results are highlighted in \colorbox{gray!30}{light gray}. \textcolor[rgb]{0.0, 0.6, 0.0}{Green font} indicates performance improvement. The best results for both per-class accuracy and overall metrics, comparing CABIN and non-CABIN methods, are shown in \textbf{bold}.}
    \label{tab:HH}
\end{table*}

